\documentclass{article}
\pdfoutput=1
\usepackage{arxiv}

\usepackage[utf8]{inputenc} 
\usepackage[T1]{fontenc}    
\usepackage{hyperref}       
\usepackage{url}            
\usepackage{booktabs}       
\usepackage{amsfonts}       
\usepackage{nicefrac}       
\usepackage{microtype}      
\usepackage{lipsum}		
\usepackage{color, graphicx}
\usepackage{natbib}
\usepackage{doi}

\usepackage{tikz}
\usepackage{amsmath}
\usepackage{pgfplots}
\usetikzlibrary{calc}
\usepackage {subfigure}
\usetikzlibrary{decorations.pathmorphing}
\usetikzlibrary {patterns,shapes,shadows,decorations.pathreplacing}
\usepackage{amssymb}
\usetikzlibrary{arrows,shadows,positioning}
\usetikzlibrary{arrows.meta}
\usepackage{multirow, makecell}
\usepackage{arydshln}
\usepackage{wrapfig}

\tikzset{pics/fake box/.style args={#1 with dimensions #2 and #3 and #4 and #5 and #6 and #7}{
    code={
       \draw[gray,ultra thin,fill=#1] (0,0,0) coordinate(-front-bottom-left) to ++ (0,#3,0) coordinate(-front-top-right) --++ (#2,0,0) coordinate(-front-top-right) --++ (0,-#3,0) coordinate(-front-bottom-right) node[right,color=black,midway,inner sep=0pt] {{\fontsize{2.0}{3}\selectfont #7}} -- cycle;
       
       \draw[gray,ultra thin,fill=#1] (0,#3,0) --++  (0,0,#4) coordinate(-back-top-left) --++ (#2,0,0)  coordinate(-back-top-right) --++ (0,0,-#4)  -- cycle; 
       
       \draw[gray,ultra thin,fill=#1!80!black] (#2,0,0) --++ (0,0,#4) coordinate(-back-bottom-right) node[right,color=black,midway,inner sep=0pt] {{\fontsize{2.0}{3}\selectfont #6}}  --++ (0,#3,0) --++ (0,0,-#4)  -- cycle;
       
       \draw[gray,ultra thin,fill=#1] (0,0,#4) --++ (0,#3,0) --++ (#2,0,0) --++ (0,-#3,0)  -- node[below,color=black,midway,inner sep=1pt] {{\fontsize{2}{3}\selectfont #5}} cycle; 
    }
  }
}

\tikzset{
        line/.style={
            draw, ->, rounded corners = 3mm,
        }
}

\tikzset{
    *|/.style={
        to path={
            (perpendicular cs: horizontal line through={(\tikztostart)},
                                 vertical line through={(\tikztotarget)})
            -- (\tikztotarget) \tikztonodes
        }
    }
}

\newcommand{\etal}{\mbox{\emph{et al.}}}

\title{Active Fire Detection in Landsat-8 Imagery: a Large-Scale Dataset and a Deep-Learning Study}
 
\author{ 
	Gabriel Henrique de Almeida Pereira
 	\\
	Department of Informatics\\
	Federal University of Technology - Paran\'{a}\\
	\texttt{pereira.gha@hotmail.com} \\
	\And
	Andr\'{e} Minoro Fusioka
 	\\
	Department of Informatics\\
	Federal University of Technology - Paran\'{a}\\
	\texttt{amfusioka@gmail.com} \\
	\AND
	Bogdan Tomoyuki Nassu \\
	Department of Informatics\\
	Federal University of Technology - Paran\'{a}\\
	\texttt{btnassu@utfpr.edu.br} \\
	\And
	Rodrigo Minetto \\
	Department of Informatics\\
	Federal University of Technology - Paran\'{a}\\
	\texttt{rminetto@utfpr.edu.br} \\
}

\date{}

\renewcommand{\headeright}{}
\renewcommand{\shorttitle}{THE FINAL VERSION OF THIS WORK IS AVAILABLE ON ISPRS Journal of Photogrammetry and Remote Sensing, Elsevier, 2021, https://doi.org/10.1016/j.isprsjprs.2021.06.002}

\hypersetup{
pdftitle={Active Fire Detection in Landsat-8 Imagery: a Large-Scale Dataset and a Deep-Learning Study},
pdfsubject={cs.AI, cs.LG},
pdfauthor={Gabriel H.~de A.~Pereira, Andr\'{e} M.~Fusioka, Bogdan T.~Nassu, Rodrigo Minetto},
pdfkeywords={active fire detection, active fire segmentation, active fire dataset, convolutional neural network, landsat-8 imagery},
}

\begin{document}
THE FINAL VERSION OF THIS WORK IS AVAILABLE ON ISPRS Journal of Photogrammetry and Remote Sensing, Elsevier, 2021, https://doi.org/10.1016/j.isprsjprs.2021.06.002

\maketitle

\begin{abstract}
Active fire detection in satellite imagery is of critical importance to the management of environmental conservation policies, supporting decision-making and law enforcement. This is a well established field, with many techniques being proposed over the years, usually based on pixel or region-level comparisons involving sensor-specific thresholds and neighborhood statistics.  In this paper, we address the problem of active fire detection using deep learning techniques. In recent years, deep learning techniques have been enjoying an enormous success in many fields, but their use for active fire detection is relatively new, with open questions and demand for datasets and architectures for evaluation. This paper addresses these issues by introducing a new large-scale dataset for active fire detection, with over 150,000 image patches (more than 200 GB of data) extracted from Landsat-8 images captured around the world in August and September 2020, containing wildfires in several locations. The dataset was split in two parts, and contains 10-band spectral images with associated outputs, produced by three well known handcrafted algorithms for active fire detection in the first part, and manually annotated masks in the second part. We also present a study on how different convolutional neural network architectures can be used to approximate these handcrafted algorithms, and how models trained on automatically segmented patches can be combined to achieve better performance than the original algorithms --- with the best combination having 87.2\% precision and 92.4\% recall on our manually annotated dataset. The proposed dataset, source codes and trained models are available on Github\footnote{\url{https://github.com/pereira-gha/activefire}}, creating opportunities for further advances in the field.

\end{abstract}


\section{Introduction}
The use of satellite imagery is of critical importance to the management of environmental conservation policies. Active fire detection is a field of research that extracts relevant knowledge from such images to support decision-making. For example, fires can be employed as a means to clear out an area after trees are cut down, giving place to pastures and agricultural land, so the presence of active fire can be directly related to accelerated deforestation in many important biomes around the world. Furthermore, there is a direct relation between biomass burning and changes to the climate and atmospheric chemistry, as stated by \cite{chuvieco19}. Active fire detection in satellite imagery is also relevant for other monitoring tasks, such as damage assessment, prevention and prediction~\footnote {https://effis.jrc.ec.europa.eu/about-effis/technical-background/rapid-damage-assessment}.

Research on the field of active fire detection focuses on providing appropriate methods for determining if a given pixel (corresponding to a georeferenced area) in a multi-spectral image corresponds to an active fire. As the satellites orbit in different altitudes and their systems are equipped with different sensors, with different wavelengths, usually such methods must be designed specifically for each system. For example, images from the Landsat-8 satellite --- the one considered in this work --- are usually processed by series of conditions such as those proposed by~\cite{SCHROEDER2016210},~\cite{MURPHY201678} or~\cite{doi:10.1080/17538947.2017.1391341}. These conditions are especially tuned for the Operational Land Imager (OLI) sensor that equips Landsat-8, as detailed in Section~\ref{section.dataset}. Other important sensors and satellites for active fire recognition are: the MODIS sensor, with 250 m to 1 km of spatial resolution per pixel, that equips the NASA Terra and Aqua satellites (orbiting 705 km height, 1-2 days of revisit); the AVHRR sensor, with 1 km of spatial resolution per pixel, that equips NOAA-18 (833 km height), NOAA-19 (870 km height) and METOP-B (827 km height) satellites, the three with a daily revisit time; the Visible Infrared Imaging Radiometer Suite (VIIRS) sensor, with 375 m of spatial resolution and onboard the joint NASA/NOAA Suomi National Polar-orbiting Partnership (Suomi NPP) and NOAA-20 satellites (824 km height), both with daily revisit time; the ABI sensor, that equips the geostationary satellite GOES-16 ($\approx$ 36,000 km height), that is able to take images each 10 minutes with spatial resolution from 500 meters to 2 km. In addition, the Sentinel-2 constellation have also been used for research on this field as they produce images similar to Landsat, with spatial resolution between 10 and 60 meters, with 5 days of revisit. 

Studies on automatic methods for detecting active fire regions date back to the 1970s, when the most suitable spectral intervals for detecting forest fires were analyzed (\cite{Kondratyev}). \cite{doi:10.1080/01431168708948657} argued that remote sensing was the only viable alternative to monitor active fire in isolated regions. In their study, they used the AVHRR sensor to detect fire activity in the Amazon area. Other active fire detection algorithms have also been developed for AVHRR (\cite{doi:10.1139/x86-171, Lee90}), some of them being used as a basis for the development of algorithms for other satellite sensors.  \cite{2002JD002331} and~\cite{ GIGLIO2003273} developed algorithms that include contextual analysis for active fire detection for the Visible and Infrared Scanner (VIRS) sensor onboard the Tropical Rainfall Measuring Mission (TRMM) satellite. \cite{98JD01644} proposed an approach for active fire detection using the MODIS sensor, which was later improved by \cite{GIGLIO2003273}. This method is still used today due to its relevance, and because the MODIS sensor is still active, and has an almost daily global coverage, being able to detect fires with size 100-300 m$^2$ (\cite{MAIER201311}). Furthermore, the MODIS detection algorithm was improved in many ways (\cite{01431160500113526,GIGLIO201631}) and served a reference for developing equations for other satellites. As an example, in 2014, \cite{SCHROEDER201485} proposed an algorithm for active fire detection for the VIIRS sensor based on the MODIS algorithm. The active fire detection algorithms mentioned above, in general, are based on comparisons to fixed threshold in certain bands and statistics from their surrounding region. 

In recent years, the tremendous advances in machine learning achieved by CNNs --- Convolutional Neural Networks --- (\cite{Lecun2015}) inspired the design of novel architectures and strategies for a wide range of applications related to active fire recognition in satellite imagery. \cite{rs11141702} used a ResNet-based classification architecture to distinguish smoke from similar patterns such as clouds, dust, haze, land and seaside, over 6,225 satellite images from the Aqua and Terra Satellites. \cite{Gargiulo2019} proposed a CNN-based architecture for image super-resolution enhancement for the specific context of wildfire. They used 10,000 image patches from the Sentinel-2 satellite, extracted from an area located in the Vesuvius volcano, in Italy. \cite{Ban2020} used temporal information from a same location, before and after wildfires, as inputs to a CNN so as to detect burned/unburned areas. The authors used in their experiments images taken by the Sentinel-1 satellite, extracting 60,000 pixels from three large wildfires in Canada and USA. \cite{8637574} tackled the imbalanced classification problem, usually present in active fire segmentation, that can negatively impact CNN performance. For their experiments they used 1,742,618 non-wildfire pixels and 105,072 wildfire pixels from a region in Alaska, collected from the Aqua and Terra Satellites. \cite{8637007} used a generative adversarial network (GAN) to synthesize missing or corrupted multispectral optical images with multitemporal data, with one application focused in wildfire detection in Sentinel-1 and Sentinel-2 images. \cite{PINTO2020260} used a deep learning approach for mapping and dating of burned areas using temporal sequences of 
VIIRS data from distinct regions from the globe. 

The use of deep learning techniques for active fire recognition is a relatively new field, which still lacks large-scale datasets and architectures for evaluation. This paper brings several contributions to the field. The first contribution is that we introduce a new public dataset for active fire recognition, built from 8,194 Landsat-8 images around the world, covering the period of August 2020, containing large wildfire events in many areas such as the Amazon region, Africa, Australia, United States, among others. The dataset contains 146,214 image patches (around 192 GB), including 10-band spectral images and associated outputs produced by three well established, handcrafted algorithms for active fire detection (\cite{SCHROEDER2016210},~\cite{MURPHY201678} and~\cite{doi:10.1080/17538947.2017.1391341}). The second contribution is a secondary dataset, also public, containing 9,044 image patches (around 12 GB) extracted from 13 Landsat-8 images captured in September 2020, along with manually annotated fire pixels, which can be employed for assessing the quality of automatic segmentation compared to a human specialist. As a third contribution, we present a study on how convolutional neural networks can be used to approximate the handcrafted fire detection algorithms mentioned above, including the possibility of saving bandwidth resources by reducing the number of analyzed spectral bands, as well as means of combining multiple fire detection outputs to produce more robust detection results. As a fourth contribution, we highlight the comparison between the fire detection algorithms mentioned above and the CNNs when tested against manual annotations from a human specialist. Finally, as a fifth contribution, all the source code of the deep learning and handcrafted algorithms employed for producing this paper are available for free access on github, all of them coded in Python language, which is a step forward for other researchers in the field towards a benchmark, opening opportunities for the study of other Landsat-8 time intervals, or the use of different satellites  --- this is a first effort in this direction, as the algorithms we analyzed here, to the best of our knowledge, do not have open source versions available. 

The rest of this paper is organized as follows. In section~\ref{section.dataset}, we describe the proposed datasets, including the procedure for generating them. In section~\ref{section.cnn}, we describe the CNN architectures for active fire detection, with experiments presented and discussed in section~\ref{section.experiments}. Finally, section~\ref{section.conclusions} presents conclusions and points to future work.



\section{Materials}~\label {section.dataset}

The Landsat program, sponsored by NASA/USGS, provides since 1972 continuous acquisition of high-quality satellite imagery of the Earth, becoming a key instrument for the continuous monitoring of the environment. As described by~\cite{ROY2014154}, the Landsat-8 satellite, which we consider in this research, orbits the Earth at an altitude of 705 km, with data segmented into scenes with $185 \times 180$ km, defined according to the second World-wide Reference System (WRS) in a 16-day revisit period. This satellite uses an Operational Land Imager (OLI) and Thermal Infrared Sensor (TIRS) sensors to acquire eleven channels of multi-spectral data $\{c_1, \dots, c_{11}\}$. The images are encoded in the TIFF format, with resolution of $\approx 7,600 \times 7,600$ pixels, with 16 bits per pixel per channel. Each multispectral pixel corresponds to 30 meters of spatial resolution (ground). 

\begin{figure}[!htb]
   \centering
   \includegraphics[width=0.9\textwidth]{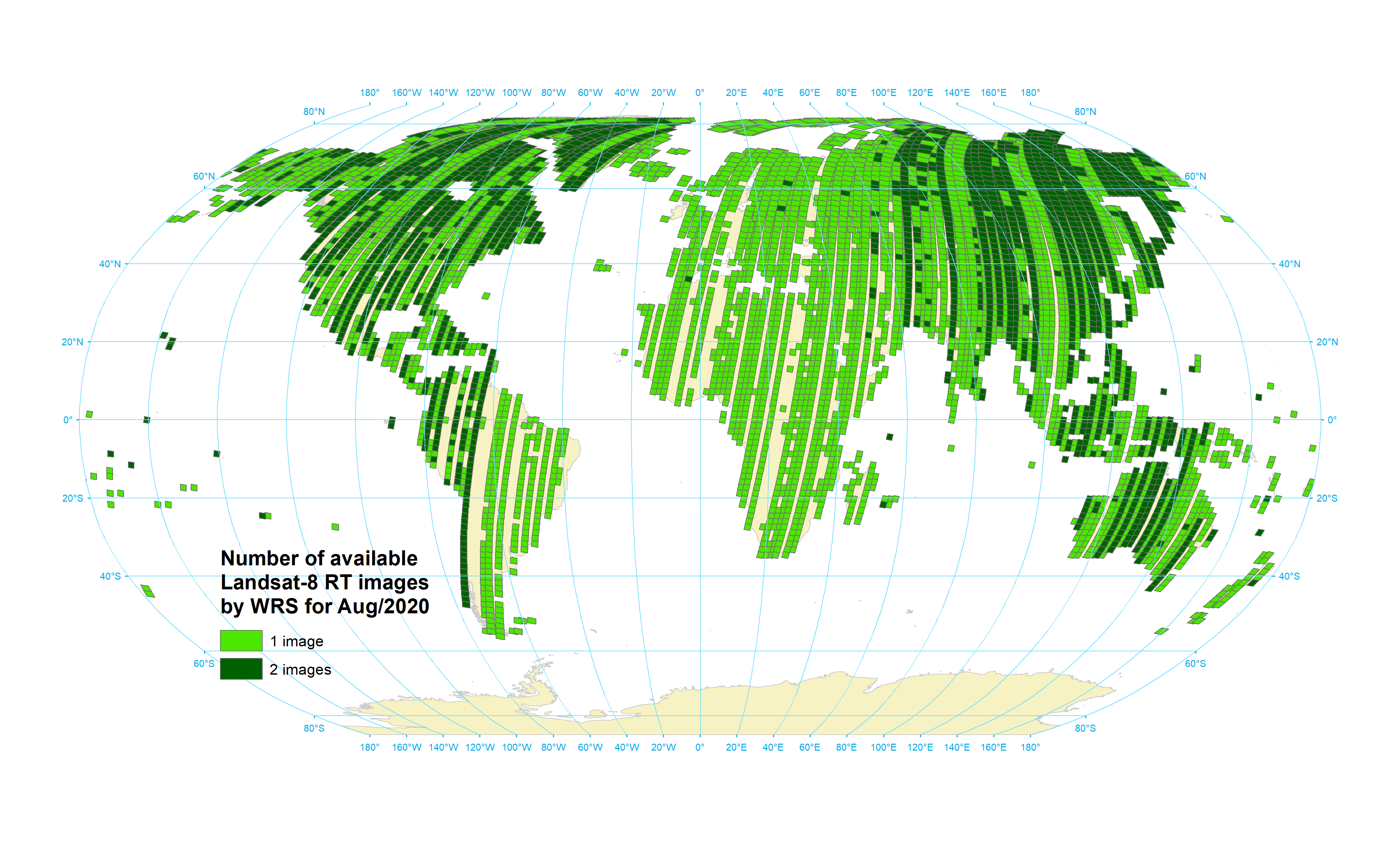}
   \caption{Landsat-8 regions used in our dataset: light green regions have one image for August 2020 and dark green regions have two images. Some regions did not have any real time images available, and are shown in yellow, along with Antartica, which was not considered in this work. Each rectangle corresponds to an Landsat-8 WRS image scene with $\approx 7,600 \times 7,600$ pixels covering a $185 \times 180$ km area.}
  \label{fig:map}
\end{figure} 

We processed all the Landsat-8 real time images available for August 2020 around the globe, excluding only the Antarctic continent, to a total of 8,194 images and 1.6 TB of data. Figure~\ref{fig:map} shows scene locations and the number of real time images available from WRS. 

The choice of this particular time slice was not random. Although the northern and southern hemispheres have different climatic behavior along the months --- especially regarding the seasons, and the occurrence of rain, snow or droughts --- fire seasons may occur around the same time in both hemispheres. \cite {ferreira20}~present a study showing the duration of fire seasons, and how they are spread along the months and across the globe. They show that August and September are the most critical months, with fire seasons occurring in all continents, except Antarctica (but including the Artic region). In 2020 this situation was not different, with several wildfire events occurring at the same time in different locations: in mid-August there were wildfires in Australia, Sibera, in the Parana Delta in Argentina, in several regions of the United States (Colorado, California, Oregon, Utah and Washington), and in the Amazon and Pantanal regions in Brazil.

\begin{figure}[!htb]
   \centering
   \includegraphics[width=0.9\textwidth]{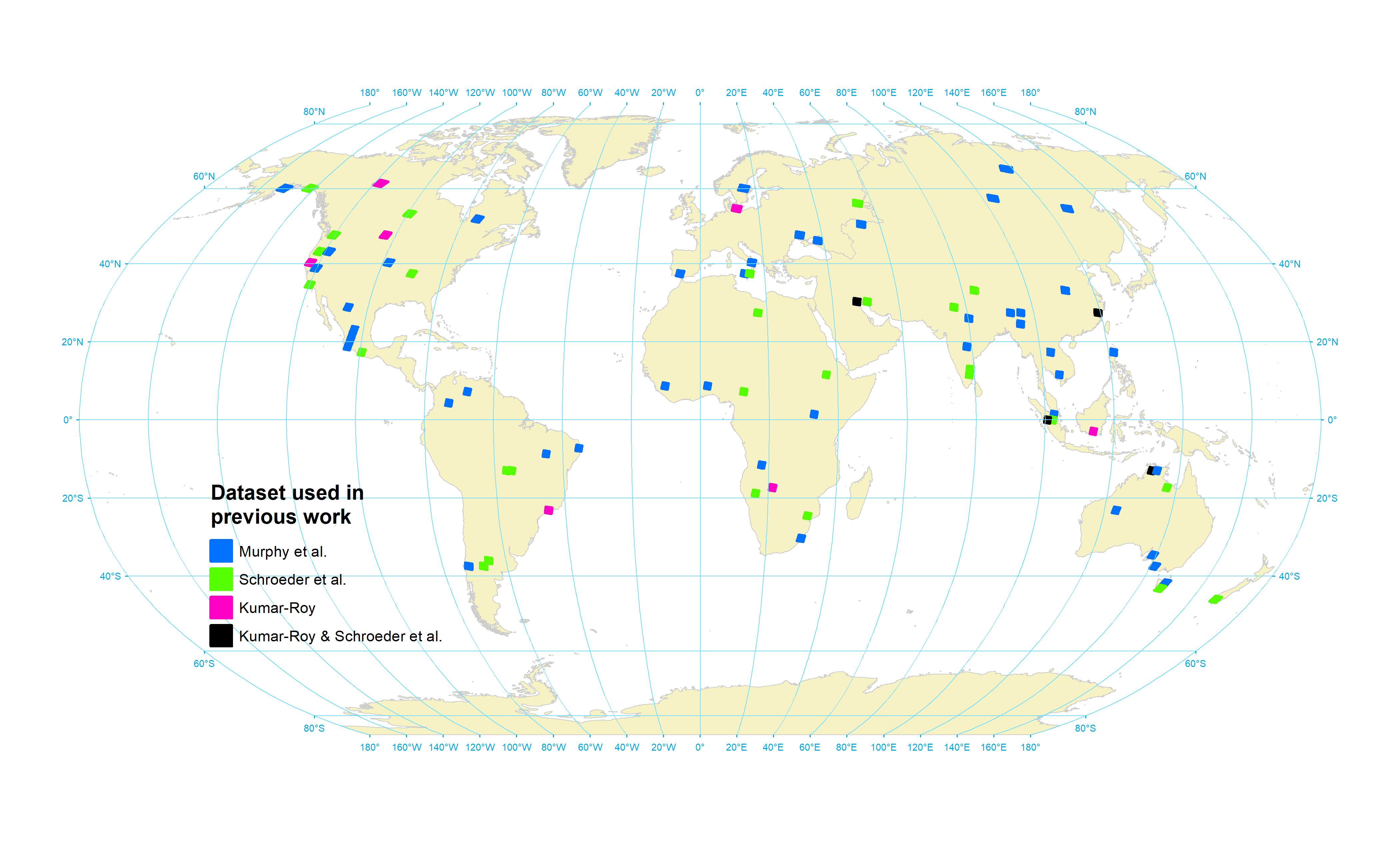}
   \caption {Locations covered in the studies from~\cite {MURPHY201678},~\cite {SCHROEDER2016210} and~\cite {doi:10.1080/17538947.2017.1391341}, which observed, respectively, 45 regions, 30 regions and 11  distinct regions.
   Multiple temporal views were used for some of these regions.
   } 
  \label{fig:mapothers}
\end{figure}

Unfortunately, it is unfeasible to manually annotate this massive amount of data in order to create masks with potential active fire pixels. However, such a large amount of data is needed for properly training deep convolutional networks. Therefore, in order to automatically generate segmentation masks, we relied on three set of conditions, well established in the field, designed by~\cite{SCHROEDER2016210},~\cite{MURPHY201678} and~\cite{doi:10.1080/17538947.2017.1391341}. Figure~\ref {fig:mapothers} shows the regions analyzed by these authors when developing and testing their sets of conditions --- note that our dataset has a more thorough coverage of the globe, as shown in Figure~\ref{fig:map}.

In the following sections, we detail the sets of conditions for active fire detection proposed by~\cite{SCHROEDER2016210},~\cite{MURPHY201678} and~\cite{doi:10.1080/17538947.2017.1391341}. For the sake of simplicity, let $\rho_i$ be in the rest of this section the reflectance of channel $c_i$. The equation to convert between Landsat-8 channels to reflectance are detailed by the USGS~\footnote{https://www.usgs.gov/land-resources/nli/landsat/using-usgs-landsat-level-1-data-product}. For the~\cite{MURPHY201678} and~\cite{doi:10.1080/17538947.2017.1391341} sets of conditions, reflectances are corrected for the solar zenith angle; while for the \cite{SCHROEDER2016210} conditions, this correction is not performed. After describing these conditions, we detail the procedure for extracting the segmentation masks, and a secondary dataset, containing manually annotated images.

\subsection{Schroeder et al. conditions}

\cite{SCHROEDER2016210}~proposed a set of conditions that uses seven Landsat-8 channels, $c_1-c_7$, to identify active fire pixels. If all conditions are satisfied the pixel is classified as true (fire), otherwise as false (non-fire). 

The first set of conditions identifies unambiguous cases:
\begin{gather*} \small
    ((R_{75} > 2.5) \;\;\textrm{and}\;\; (\rho_7 - \rho_5 > 0.3) \;\;\textrm{and}\;\; (\rho_7 > 0.5)) \textrm { or} \\
    ((\rho_6 > 0.8) \;\;\textrm{and}\;\; (\rho_1 < 0.2) \;\;\textrm{and}\;\; (\rho_5 > 0.4 \;\;\textrm{or}\;\ \rho_7 < 0.1)) 
\end{gather*}
where $R_{ij}$ is the ratio between the reflectance in channels $i$ and $j$ ($\rho_i / \rho_j$).

Then, by using a neighborhood of $61 \times 61$ pixels, centered at each fire pixel candidate, the above conditions are relaxed so as to consider the region context, namely
\begin{gather*} \small 
    (R_{75} > 1.8) \;\textrm{and}\; (\rho_7 - \rho_5 > 0.17)  \;\textrm{and}\; 
     (R_{76} > 1.6) \;\textrm{and}\; \\ 
    (R_{75} > \mu_{R_{75}}\textrm{+}\max(3\sigma_{R_{75}},\textrm{0.8})) \;\textrm{and}\;  (\rho_7 > \mu_{\rho_{7}}\textrm{+}\max(3\sigma_{\rho_{7}},\textrm{0.08}))
\end{gather*}
where $\mu$ and $\sigma$ denote the mean and standard deviation of the pixels in a $61 \times 61$ window pixel neighborhood. These means and standard deviations are computed excluding fire and water pixels, the latter being classified by:
\begin{gather*} \small
    (\rho_4 > \rho_5) \;\textrm{and}\; (\rho_5 > \rho_6) \;\textrm{and}\; (\rho_6 > \rho_7) \;\textrm{and}\; (\rho_1 - \rho_7 < 0.2) \\
    \;\textrm{and}\; ((\rho_3 > \rho_2) \;\textrm{or}\; ((\rho_1 > \rho_2) \;\textrm{and}\; (\rho_2 > \rho_3) \;\textrm{and}\; (\rho_3 > \rho_4)))
\end{gather*}

For further details about the above conditions refer to the work of~\cite{SCHROEDER2016210}. 

\subsection{Murphy et al. conditions}

\cite{MURPHY201678}~proposed a set of conditions that is non-contextual, in the sense that it does not rely on statistics computed from each pixel's neighborhood. It is based on channels $c_5$, $c_6$ and $c_7$.

Unambiguous active fire pixels are first identified by:
\begin{gather*} \small
    (R_{76} \geq 1.4) \;\;\textrm{and}\;\; (R_{75} \geq 1.4) \;\;\textrm{and}\;\; (\rho_7 \geq 0.15)
\end{gather*}
where $R_{ij}$ is the ratio between the reflectance in channels $i$ and $j$. Potential fires that are in the immediate neighborhood of an unambiguous fire (in a $3 \times 3$ window), are also classified as fires. Potential fires are those that satisfy: 
\begin{gather*} \small
    ((R_{65} \geq 2) \;\;\textrm{and}\;\; (\rho_6 \geq 0.5))  \;\;\textrm{or}\;\; ((\rho_7 \;\;\textrm{is saturated}) \;\;\textrm{or}\;\; (\rho_6 \;\;\textrm{is saturated} ) )
\end{gather*}

Criteria for identifying saturated bands are defined by the USGS~\footnote{https://www.usgs.gov/land-resources/nli/landsat/landsat-collection-1-level-1-quality-assessment-band}.

\subsection{Kumar-Roy conditions}

The conditions proposed by~\cite{doi:10.1080/17538947.2017.1391341} are based on channels $c_2-c_7$. This work is more recent than the others, and builds upon some of the insights behind previous work. Unambiguous fire pixels are identified as those that satisfy
\begin{gather*} \small
    \rho_4 \leq 0.53 \; \rho_7 - 0.214
\end{gather*}

They also take as unambiguous fire pixels those that are in the immediate 8-pixel vicinity of a fire pixel classified by the above condition, and which satisfy the more relaxed condition
\begin{gather*} \small
    \rho_4 \leq 0.35 \; \rho_6 - 0.044
\end{gather*}

Potential fire pixels are identified by
\begin{gather*} \small
    ( (\rho_4 \leq 0.53 \; \rho_7 - 0.125)  \;\;\textrm{or}\;\; (\rho_6 \leq 1.08 \; \rho_7 - 0.048) )
\end{gather*}
and are only kept if they satisfy the contextual test
\begin{gather*} \small
    (R_{75} > \mu_{R_{75}}\textrm{+}\max(3\sigma_{R_{75}},\textrm{0.8})) \;\textrm{and}\;  (\rho_7 > \mu_{\rho_{7}}\textrm{+}\max(3\sigma_{\rho_{7}},\textrm{0.08}))
\end{gather*}

where $R_{75}$ is the ratio between the reflectance in channels 7 and 5 ($\rho_7 / \rho_5$), and $\mu$ and $\sigma$ denote the mean and standard deviation of the pixels in a neighborhood around each candidate fire pixel. This contextual test is the same used by the~\cite{SCHROEDER2016210} conditions, with two differences. The first difference is the condition used to detect water pixels

\begin{gather*} \small
    (\rho_2 \geq \rho_3 \geq \rho_4 \geq \rho_5)
\end{gather*}

The second difference is that the region size is not fixed: we test progressively larger neighborhoods ($5 \times 5$, $7 \times 7$, $9 \times 9$, and so on, up to maximum of $61 \times 61$ pixels), stopping when the region contains at least 25\% of pixels not classified as unambiguous or potential fires, or as water --- these pixels are excluded when computing the means and standard deviations. One point that is not addressed in the original paper is how to proceed when the equations detect a large area containing potential fire pixels --- in these cases, even with a $61 \times 61$ neighborhood there are candidate fire pixels mostly surrounded by excluded pixels, and the contextual test becomes unreliable. We considered such cases as non-fires, since doing otherwise made the algorithm highly prone to false detections.

\subsection{Segmentation masks}

\begin{figure}[!htb]
   \hspace{-15pt}
   \includegraphics[width=1.05\textwidth]{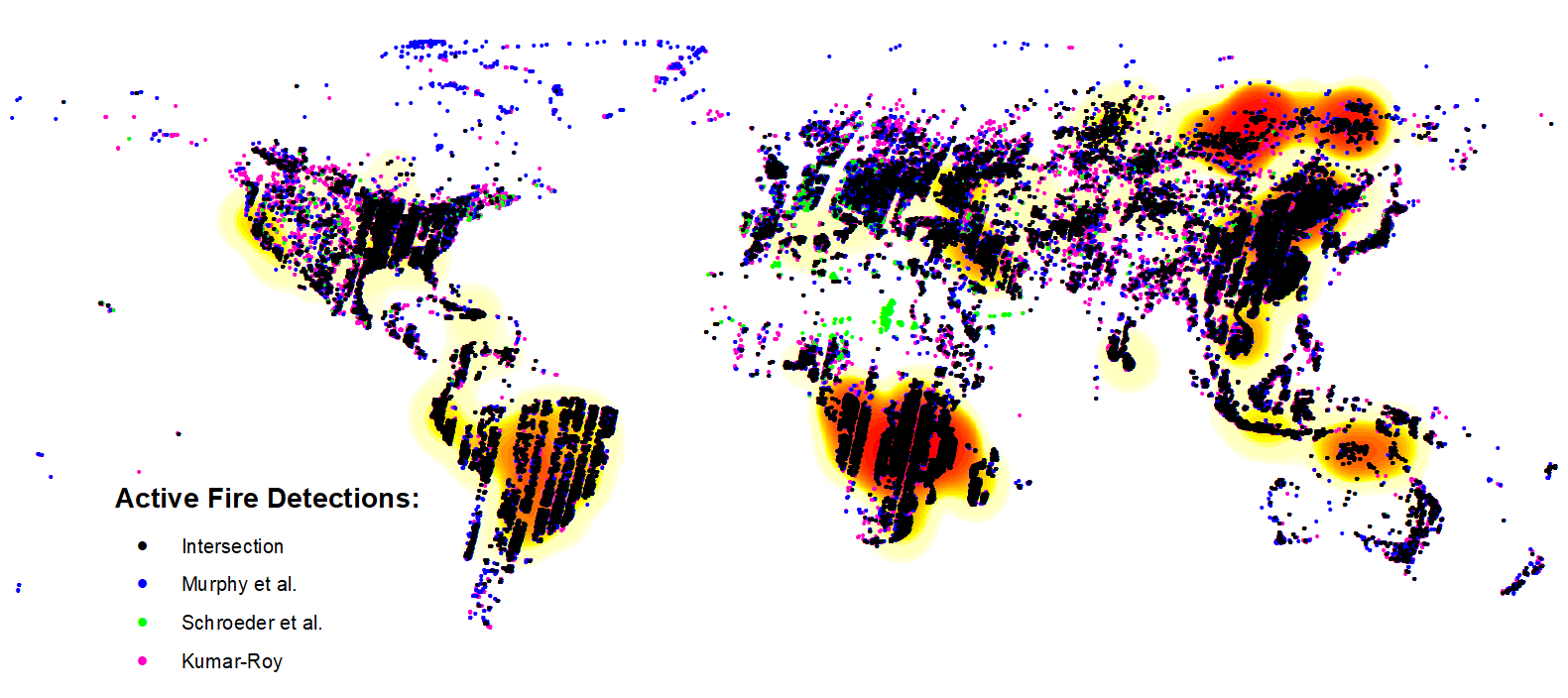}
   \caption{Fire incidence detection and heat map around the globe for August 2020 considering the outputs of~\cite{SCHROEDER2016210},~\cite{MURPHY201678} and~\cite{doi:10.1080/17538947.2017.1391341}. Black dots represent wildfire locations simultaneously detected by the three algorithms.}
  \label{fig:heatmap}
\end{figure}

In order to create segmentation masks, we used
the conditions from~\cite{SCHROEDER2016210},~\cite{MURPHY201678} and~\cite{doi:10.1080/17538947.2017.1391341} over the original scenes with $\approx 7,600 \times 7,600$ pixels. Figure~\ref{fig:heatmap} shows the locations were fires have occurred in our dataset for each set of conditions.

\begin{figure}[!htb]
   \centering
   \includegraphics[width=0.85\textwidth]{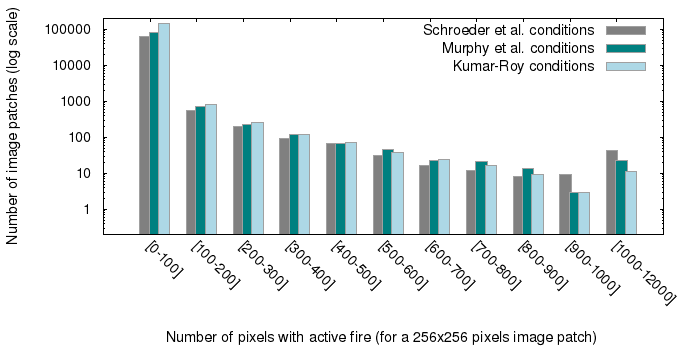}
   \caption{Distribution of image patches per fire pixel count.
   The total number of image patches (with $256 \times 256$ pixel size) for the Schroeder et al.~conditions is 62,385, for Murphy et al.~conditions is 82,036 and 143,133 for the Kumar-Roy conditions.}
  \label{fig:histogram}
\end{figure}

This dataset was then cropped into image patches with $256 \times 256$ pixels without overlap. The number of fire pixels per image patch for each set of conditions is shown in the histogram in Figure~\ref{fig:histogram}. It can be seen that most patches have a small number of fire pixels. 

Our dataset is publicly available, and contains 146,214 image patches, $256 \times 256$ pixels each, in a total of 192 GB, along with their corresponding segmentation masks produced by the three sets of conditions discussed above (approximately 22 GB). The patches are 10-band, 16-bit TIFF images, with channels $c_1, \dots, c_7$ and $c_9, \dots, c_{11}$. The panchromatic channel $c_8$, with 15 meters of spatial resolution, will not be considered in this work.

\begin{figure}[!htb]
   \centering
   \setlength{\tabcolsep}{2pt}
   \begin{tabular}{cccc}
   {\scriptsize Input} &
   {\scriptsize Schroeder et al.}&
   {\scriptsize Murphy et al.}&
   {\scriptsize Kumar-Roy}\\
   \includegraphics[width=0.22\textwidth]{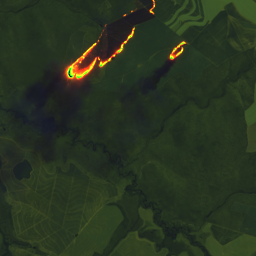} &
   \includegraphics[width=0.22\textwidth]{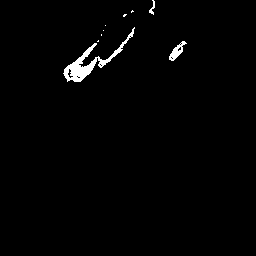}&
   \includegraphics[width=0.22\textwidth]{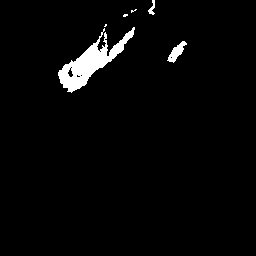}&
   \includegraphics[width=0.22\textwidth]{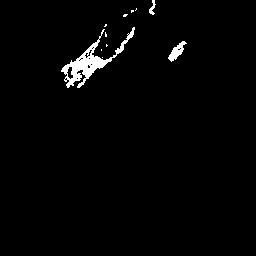}\\
   \includegraphics[width=0.22\textwidth]{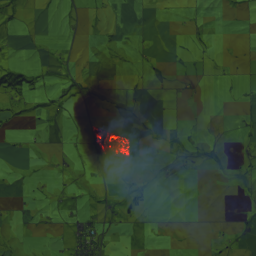} &
   \includegraphics[width=0.22\textwidth]{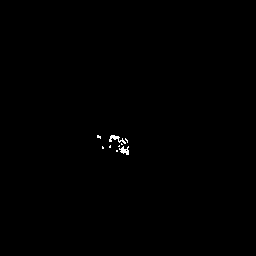}&
   \includegraphics[width=0.22\textwidth]{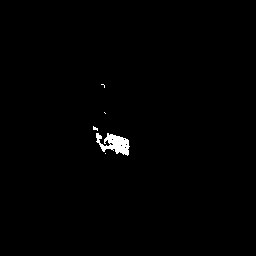}&
   \includegraphics[width=0.22\textwidth]{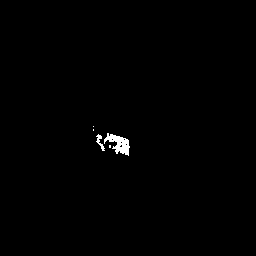}\\
   \includegraphics[width=0.22\textwidth]{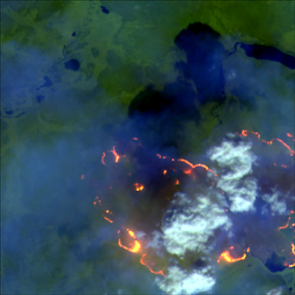} &
   \includegraphics[width=0.22\textwidth]{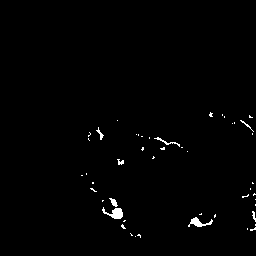}&
   \includegraphics[width=0.22\textwidth]{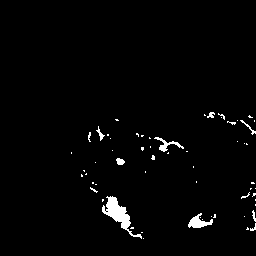}&
   \includegraphics[width=0.22\textwidth]{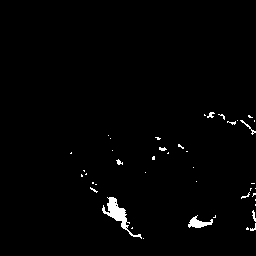}\\
   \end{tabular}
   \caption{Image patches and the corresponding segmentation masks, extracted from the proposed dataset. For display purposes we used channels $c_7$, $c_6$ and $c_2$ to compose the RGB bands, however, the original patches contain 10 bands.}
  \label{fig:detections_equation}
\end{figure}

Figure~\ref{fig:detections_equation} shows some image samples with active fire and the corresponding segmentation masks for the three set of conditions. It can be seen that, although the three algorithms generally agree on the presence of fire on a certain level, the pixelwise segmentation tends to differ between them. The~\cite{MURPHY201678} conditions seem more sensitive to low level intensities (ember), while all the algorithms have problems with very high intensities (which appear with a greenish tinge in the RGB visualization). These differences will be further discussed in Section \ref {section.experiments}.

We believe that this dataset can prove useful and relevant for other researchers working on active fire detection, since it provides the Landsat-8 data in a friendly format that can be directly used by existing machine learning tools, and provides a challenging target for benchmarking, with samples covering a large variety of scenarios, including desert, rain forest, agricultural land, water, snow, clouds, cities, mountains, etc.

\subsection{Manually annotated dataset}

\begin{figure}[!htb]
   \centering
   \includegraphics[width=0.9\textwidth]{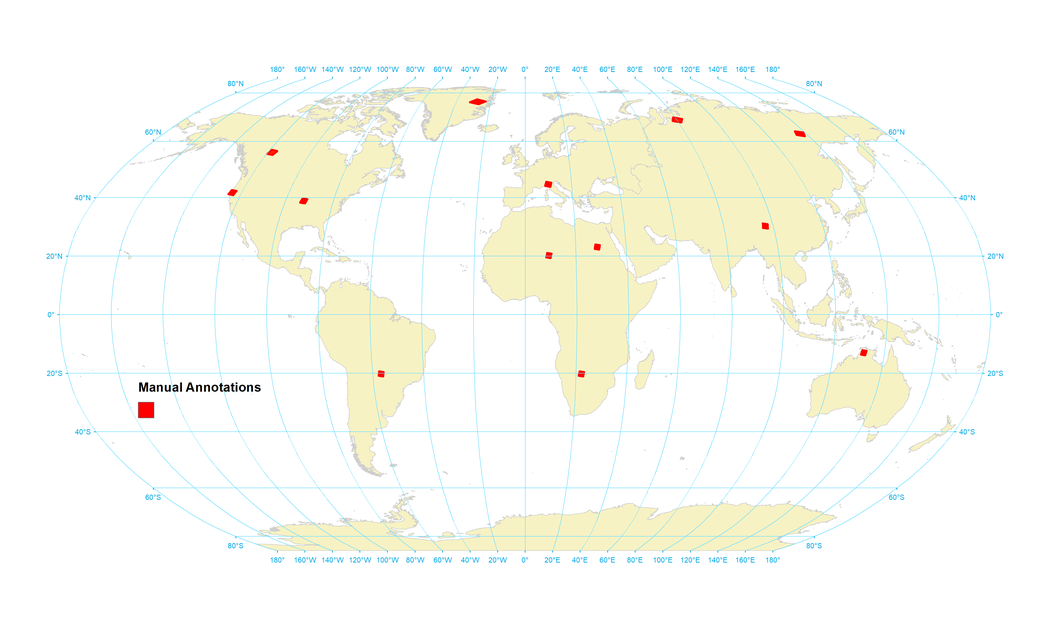}
   \caption{Active fire manual annotations: the active fire present in 13 Landsat-8 images (depicted by red squares) from September of 2020 --- with $\approx 7,600 \times 7,600$ pixels and selected to represent distinct regions of the planet --- were manually annotated by the authors. These images and reference masks are also available as a dataset.}
  \label{fig:manual}
\end{figure}

To further validate our trained models, we selected 13 Landsat-8 images, with $\approx 7,600 \times 7,600$ pixels, which were split into 9,044 image patches with $256 \times 256$ pixels each, without overlap, in a total of 12 GB, captured in September 2020, and manually annotated them. Figure~\ref {fig:manual} shows the locations of these images.

\begin{figure}[!htb]
	\centering
	\includegraphics[width=0.9\textwidth]{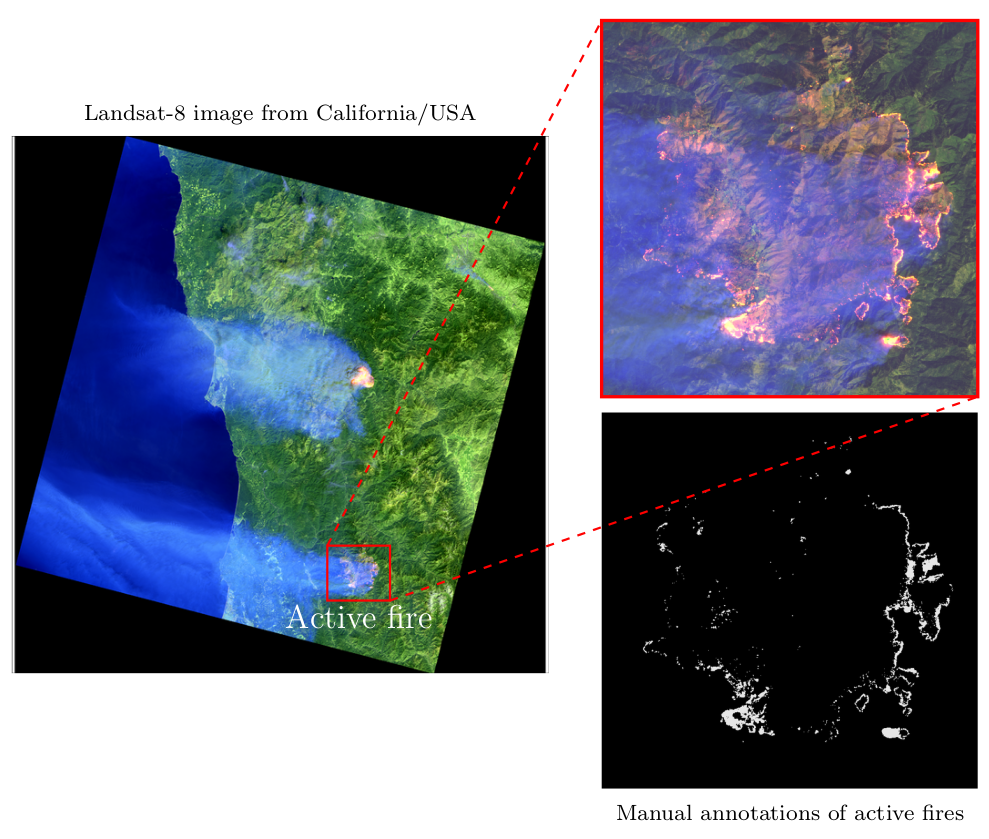}
   \caption{Manual annotation example: on the left, a Landsat-8 image from California, USA --- reference code LC08\_L1TP\_046031\_20200908\_20200908\_01\_RT, 
   WRS 046/031, from September 8th, 2020. This image has many wildfire regions, with a total of 55,386 active fire pixels being identified. Highlighted in red, one of the two major wildfire areas (top right) with an extension of $22\times22$ km, with approximately 10,000 active fire pixels; the corresponding manual annotations are also shown (bottom right).}
  \label{fig:manual_sample}
\end{figure}

It can be seen that these images are well spread spatially around the globe, covering the main continents, as well as the different features that may exist in these regions. There are active fires in 10 of these images, with different cloud cover and fire properties --- some have many fire pixels, some have only a few; some have large fires, some have small isolated fire pixels; with the fire intensity also varying. The remaining 3 images have no active fires, but cover scenarios that proved challenging for at least one of the sets of conditions, which produced false positives: one from the Sahara desert, where the~\cite {SCHROEDER2016210} conditions detected fires, one from Greenland, where the~\cite{MURPHY201678} and~\cite{doi:10.1080/17538947.2017.1391341} conditions detected fires, and one from a city (Milan), since all the algorithms seem to produce false detections inside large urban settlements. These problems can be observed in Figure \ref {fig:heatmap}, which refers to a different time slice, but also has some detections in these regions. Figure~\ref {fig:manual_sample} shows an example of manually annotated image.

The purpose of evaluating the learned models against these images is having a second reference for validating them, avoiding some potential bias that might arise from training them on the outputs from the~\cite{SCHROEDER2016210},~\cite{MURPHY201678} and~\cite{doi:10.1080/17538947.2017.1391341} algorithms. For this reason, in this work, the manually annotated images will be exclusively for testing the models, but not for training them.


\section{Methods}
\label {section.cnn}

Convolutional neural networks (CNNs)(\cite{10.5555/3086952}) are a type of feedforward neural network architecture that, in recent years, have been enjoying a lot of attention, due to their success in a large variety of tasks, many of them related to image processing (\cite{GARCIAGARCIA201841,9044873}), analysis (\cite{7820963,LITJENS201760}) and recognition (\cite{PAOLETTI2019279,YAO201914}). This popularization is mainly because to advances both in processing capabilities --- particularly the use of Graphic Processing Units (GPUs) for massively parallel computations --- and algorithms and techniques, which led to the ability of learning deeper and more complex models. The reviews by~\cite{8113128}, ~\cite{MA2019166} and~\cite{YUAN2020111716} discuss 
the major breakthroughs of deep learning for the remote sensing field, including 
main challenges, and the current state-of-the-art in environmental parameter retrieval, data fusion and downscaling, image registration, scene classification, object detection, land use and land cover classification and region segmentation.

A deep convolutional network is composed by a series of layers, each layer applying to the output of the previous layer a number of filters via the linear convolution operation, followed by some kind of activation function that introduces non-linearities~(\cite{10.5555/3086952}). Deeper layers combine features extracted by previous layers, in such a way that the network is capable of encoding progressively more complex concepts (\cite{Lecun2015}). The training procedure for a CNN consists of discovering weights (coefficients and biases for the filters), which allow the input data to be transformed so as to approximate the desired output. In this paper, we consider only the so called supervised learning~(\cite{10.5555/1841234}), in which the network receives a number of labeled samples --- inputs and their associated outputs --- and iteratively adjusts its weights via a backpropagation algorithm~(\cite{Rumelhart:1986we}) to make its output for a given input more and more similar to the presented sample output. Although CNNs are the current state of the art technique for a number of machine learning and computer vision tasks, there are several concerns that must be kept in mind when working with them. In this paper, we do not focus on issues directly related to challenges faced by CNNs in a mathematical or algorithmic level, but we do highlight that complex architectures demand a high computational power to be trained (i.e.~whenever possible, it is desirable to have smaller and simpler architectures); and that a proper model must be trained with a large amount of unbiased data, something that raises concerns over both the required infrastructure needed for transmission/storage and the availability of good quality public datasets.

The design of robust and highly optimized CNN architectures for active fire detection is not the main focus of this work. Nevertheless, it is important to demonstrate that a CNN trained on the proposed dataset is able to produce results similar to those obtained by conventional algorithms such as those from~\cite{SCHROEDER2016210},~\cite{MURPHY201678} and~\cite{doi:10.1080/17538947.2017.1391341}. Our tests were based on networks derived from the U-Net architecture from \cite{RFB15a}, a very popular architecture for image segmentation tasks. U-Net is a fully convolutional network, with two symmetric halves, the first with pooling operations that decrease the data resolution and the second with upsampling operations that restore the data to its original resolution. The first half extracts basic features and context, and the second half allows for a precise pixel-level localization of each feature, with skip connections between the two halves being used to combine the features from a layer with more abstract features from deeper layers.

\begin{figure}[!htb]
   \includegraphics[width=1.0\textwidth]{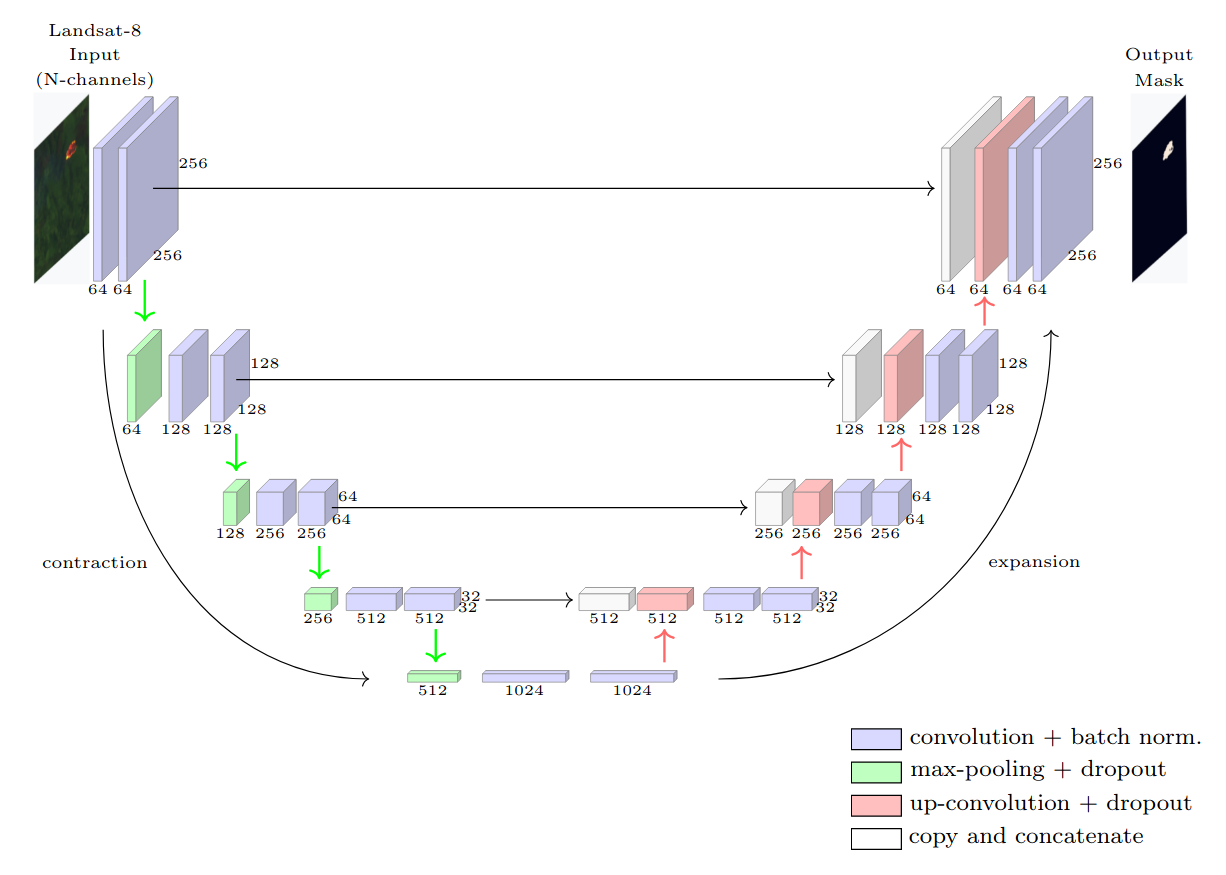}
 	\caption{U-Net architecture for image segmentation.}
  \label{fig:cnn_unet}
\end{figure}

Figure~\ref{fig:cnn_unet} shows the basic U-Net architecture employed in our tests. It is essentially the same concept from the original U-Net, but our implementation adds batch normalization after each convolutional layer, and contains additional dropout layers, which help avoiding problems with overfitting~(\cite{10.5555/3086952}) and is related to how the error propagates during training. 

We considered 3 variations of the U-Net architecture. The first one, called hereafter U-Net (10c), is the basic U-Net architecture, taking as input a 10-channel image containing all the Landsat-8 image bands. The second architecture, called hereafter U-Net (3c), keeps the same structure, but replacing the input with a 3-channel image containing only bands $c_7, c_6$, and $c_2$ --- the rationale behind this decision is checking whether it is possible to obtain a good enough approximation of the results while relying on a reduced number of bands, something that may be useful for reducing bandwidth, memory usage and storage space, e.g. it is possible to have a considerable economy of bandwidth resources when processing the images in Amazon Web Services (AWS), where the original Landsat-8 images are hosted for free download and where is possible to download only selected bands.

\begin{figure}[!htb]
   \centering
   \includegraphics[width=0.32\textwidth]{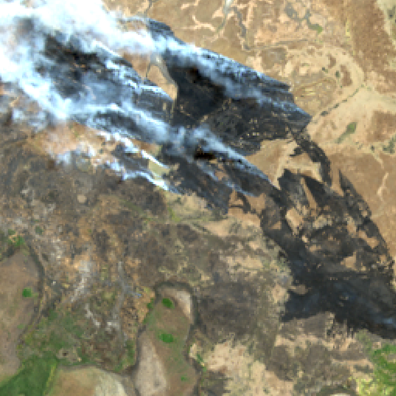} \hspace{15pt}
   \includegraphics[width=0.32\textwidth]{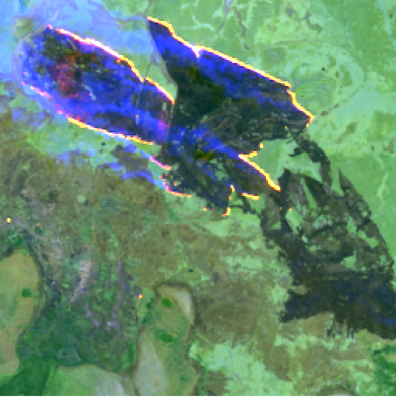}
   \caption{Landsat-8 channel composition: on the left the composition of channels $c_4$, $c_3$ and $c_2$; on the right the composition of channels $c_7$, $c_6$ and $c_2$, as we are using in our 3-channel CNNs. Note that only the smoke is visible in the left image, while the fire appears clearly in the right image.}
  \label{fig:channels_difference}
\end{figure}

Figure~\ref{fig:channels_difference} shows the difference between two images from the same scene but using different combinations of Landsat-8 channels. 

The third architecture, called hereafter U-Net-Light (3c), is a reduced version of the 3-channel U-Net, with the number of filters per layer being divided by 4 (i.e.~16 filters in the first layer, 32 in the second, etc.). We call this a ``light'' version of the original U-Net network. Relevant parameters for these architectures are shown in Table~\ref{tab:cnn_parameters}.
\begin{table}[!htb] \footnotesize
   \setlength{\tabcolsep}{6pt}
   \renewcommand{\arraystretch}{1.1}
   \centering
   \caption{U-Net-based architectures for active fire recognition.}
   \label{tab:cnn_parameters}
   \begin{tabular}{ccccc}
    \hline \hline
    & \# of input &  \# of trainable  &  model size         & inference time  \\ 
    & channels & parameters   &    (MB)        & per patch (ms) \\ \hline
   U-Net (10c)  & 10    &  34,529,153  &  132    & 36,8 \\
   U-Net (3c)   & 3    &  34,525,121   &  132    & 36 \\
   U-Net-Light (3c) & 3 &  2,161,649   &  8,5    & 25,5 \\ \hline
   \end{tabular}
\end{table}

All the variations have as output a 1-channel binary image, with $256 \times 256$ pixels, where 1 and 0 represent, respectively, fire and non-fire locations. To obtain these binary outputs, the CNN outputs are thresholded so that any pixel with value above 0.25 is set to 1 (this threshold was empirically defined, after an initial test run on a small fraction of the dataset).

\begin{figure}[!htb]
\centering
\subfigure[]{
  \begin{tikzpicture}[]
   \tikzset{blockN/.style={draw, rectangle, text centered, drop shadow, fill=blue!10!white, text width=3.0cm, minimum height=0.5cm}}
   
   \tikzset{blockM/.style={draw, rectangle, text centered, drop shadow, fill=white, text width=3.0cm, minimum height=0.5cm}}

   \fill[gray!10!white] (-0.8,-0.47) rectangle (7.8,1.68);

   \path[->](0.5,1.4) node[black]  {\scriptsize Input (N-band)};
   \node[inner sep=0pt] (img) at (0.5,0.5) { \includegraphics[width=1.5cm,height=1.5cm]{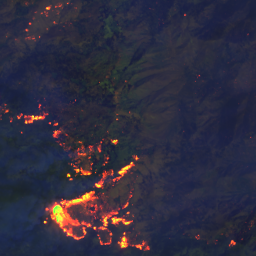}};

   \path[->](6.5,1.4) node[black]  {\scriptsize Mask};
   \node[inner sep=0pt] (mask) at (6.5,0.5) { \includegraphics[width=1.5cm,height=1.5cm]{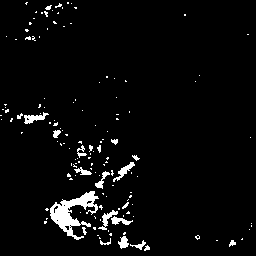}};
   
   \path[->](3.5,0.1) node[blockM] (unet) {
      \scriptsize U-Net training
   };
   
   \path[->](3.5,0.9) node[blockN] (kr) {
      \scriptsize Schroeder conditions
   };
   
   \draw[->] (img.east |- kr) -- (kr);
   \draw[->] (img.east |- unet) -- (unet);
   \draw[<-] (mask.west |- kr) -- (kr);
   \draw[->] (mask.west |- unet) -- (unet);
  \end{tikzpicture}
}

\subfigure[]{
  \begin{tikzpicture}[]
   \tikzset{blockN/.style={draw, rectangle, text centered, drop shadow, fill=blue!60!green!20, text width=3.0cm, minimum height=0.5cm}}
   
   \tikzset{blockM/.style={draw, rectangle, text centered, drop shadow, fill=white, text width=3.0cm, minimum height=0.5cm}}

   \fill[gray!10!white] (-0.8,-0.47) rectangle (7.8,1.68);

   \path[->](0.5,1.4) node[black]  {\scriptsize Input (N-band)};
   \node[inner sep=0pt] (img) at (0.5,0.5) { \includegraphics[width=1.5cm,height=1.5cm]{./figures/scheme/LC08_L1TP_046031_20200908_20200908_01_RT_p00615_762.png}};
  
   \path[->](6.5,1.4) node[black]  {\scriptsize Mask};
   \node[inner sep=0pt] (mask) at (6.5,0.5) { \includegraphics[width=1.5cm,height=1.5cm]{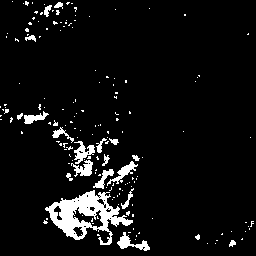}};
   
   \path[->](3.5,0.1) node[blockM] (unet) {
      \scriptsize U-Net training
   };
   
   \path[->](3.5,0.9) node[blockN] (kr) {
      \scriptsize Murphy conditions
   };
   
   \draw[->] (img.east |- kr) -- (kr);
   \draw[->] (img.east |- unet) -- (unet);
   \draw[<-] (mask.west |- kr) -- (kr);
   \draw[->] (mask.west |- unet) -- (unet);
  \end{tikzpicture}
}

\subfigure[]{
  \begin{tikzpicture}[]
   \tikzset{blockN/.style={draw, rectangle, text centered, drop shadow, fill=red!60!green!30, text width=3.0cm, minimum height=0.5cm}}
   
   \tikzset{blockM/.style={draw, rectangle, text centered, drop shadow, fill=white, text width=3.0cm, minimum height=0.5cm}}

   \fill[gray!10!white] (-0.8,-0.47) rectangle (7.8,1.68);

   \path[->](0.5,1.4) node[black]  {\scriptsize Input (N-band)};
   \node[inner sep=0pt] (img) at (0.5,0.5) { \includegraphics[width=1.5cm,height=1.5cm]{./figures/scheme/LC08_L1TP_046031_20200908_20200908_01_RT_p00615_762.png}};
  
   \path[->](6.5,1.4) node[black]  {\scriptsize Mask};
   \node[inner sep=0pt] (mask) at (6.5,0.5) { \includegraphics[width=1.5cm,height=1.5cm]{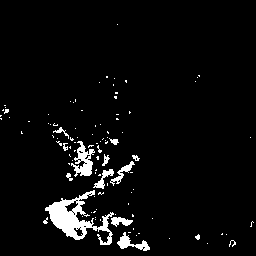}};
   
   \path[->](3.5,0.1) node[blockM] (unet) {
      \scriptsize U-Net training
   };
   
   \path[->](3.5,0.9) node[blockN] (kr) {
      \scriptsize Kumar-Roy conditions
   };
   
   \draw[->] (img.east |- kr) -- (kr);
   \draw[->] (img.east |- unet) -- (unet);
   \draw[<-] (mask.west |- kr) -- (kr);
   \draw[->] (mask.west |- unet) -- (unet);
  \end{tikzpicture}
} 
\subfigure[]{
  \begin{tikzpicture}[]
   \tikzset{blockN1/.style={draw, rectangle, text centered, drop shadow, fill=blue!10!white, text width=3.0cm, minimum height=0.5cm}}
   
   \tikzset{blockN2/.style={draw, rectangle, text centered, drop shadow, fill=blue!60!green!20, text width=3.0cm, minimum height=0.5cm}}
   
   \tikzset{blockN3/.style={draw, rectangle, text centered, drop shadow, fill=red!60!green!30, text width=3.0cm, minimum height=0.5cm}}
   
   \tikzset{blockM/.style={draw, rectangle, text centered, drop shadow, fill=white, text width=3.0cm, minimum height=0.5cm}}
   
   \tikzset{blockUI/.style={draw, rectangle, text centered, drop shadow, fill=red!40!white!30, text width=1.8cm, minimum height=0.5cm}}

   \fill[gray!10!white] (-1.05,-1.05) rectangle (10.5,2.1);

   \path[->](0.3,1.4) node[black]  {\scriptsize Input (N-band)};
   \node[inner sep=0pt] (img) at (0.3,0.5) { \includegraphics[width=1.5cm,height=1.5cm]{./figures/scheme/LC08_L1TP_046031_20200908_20200908_01_RT_p00615_762.png}};
  
   \path[->](9.3,1.8) node[black]  {\scriptsize Mask};
   \node[inner sep=0pt] (mask) at (9.3,0.9) { \includegraphics[width=1.5cm,height=1.5cm]{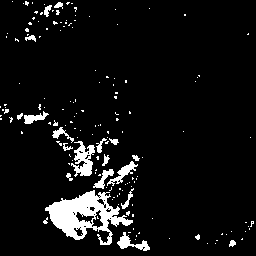}};
   
   \path[->](3.5,-0.5) node[blockM] (unet) {
      \scriptsize U-Net training
   };
   
   \path[->](3.5,1.6) node[blockN1] (sc) {
      \scriptsize Schroeder conditions
   };
   
   \path[->](3.5,0.9) node[blockN2] (mc) {
      \scriptsize Murphy conditions
   };
   
   \path[->](3.5,0.2) node[blockN3] (kc) {
      \scriptsize Kumar-Roy conditions
   };
   
   \path[->](6.9,0.9) node[blockUI] (uori) {
      \scriptsize Intersection\\[-0.4mm]or\\[-2.2mm]Voting
   };
   
   \draw[] (img) -- (1.45,0.5);
   \draw[->] (1.45,0.5) |- (sc);
   \draw[->] (1.45,0.5) |- (mc);
   \draw[->] (1.45,0.5) |- (kc);
   \draw[->] (1.45,0.5) |- (unet);
   
   \draw[] (sc) -| (5.5,0.9);
   \draw[] (mc) -| (5.5,0.9);
   \draw[] (kc) -| (5.5,0.9);
   \draw[->] (5.5,0.9) -- (uori);
   
   \draw[->] (uori) -- (mask);
   \draw[->] (mask.south) |- (unet);
  \end{tikzpicture}
}
\caption{Different situations for training and testing the deep convolutional neural networks. Each of the 3 network architectures (U-Net (10c),  U-Net (3c) and  U-Net-Light (3c)) was trained and tested to approximate the outputs of 5 different sets of conditions: the 3 algorithms for active fire detection \cite{SCHROEDER2016210},~\cite{MURPHY201678} and~\cite{doi:10.1080/17538947.2017.1391341}, as shown in (a), (b) and (c), as well as their intersection and a ``best-of-three'' voting, as shown in (d).}
\label{fig:situations}
\end{figure}

Each of the 3 CNN architectures was trained and tested to approximate 5 different situations: each of the 3 considered sets of conditions, as well as their intersection and a ``best-of-three'' voting (i.e.~a pixel in the mask is set as active fire if at least two sets of conditions agree that it is a fire pixel). This adds up to a total of 15 tested scenarios, as illustrated in Figure~\ref {fig:situations}.

\section{Results and Discussion}~\label{section.experiments}

The proposed CNN architectures were implemented and tested, to verify their ability to approximate the results from the sets of conditions from~\cite{SCHROEDER2016210}, \cite{MURPHY201678} and \cite{doi:10.1080/17538947.2017.1391341}. We also evaluated the performance of the CNNs and sets of conditions on a set of manually annotated images.

The image patches from the dataset were randomly divided in 3 sets, for training, validation and test, containing, respectively, 40\%, 10\% and 50\% of the samples. The experiments were carried out on an Intel Core i8, 64 GB of RAM, running Linux, with a Titan Xp (12 GB) GPU. The implementation was in the Python language, using the TensorFlow\footnote {www.tensorflow.org} library. We trained each architecture using Adam optimization (learning rate of 0.001), a batch size of 16, and binary cross entropy as the loss function, for 50 epochs, or until the loss on the validation set was not improved for at least 5 epochs.

\subsection{Evaluation metrics}

We evaluated the architectures according to 
the primary metrics used in the main semantic segmentation challenges such as PASCAL VOC~(\cite{10.1007/s11263-009-0275-4}), KITTI~(\cite{6728473}), and COCO\footnote{http://cocodataset.org}(~\cite{10.1007/978-3-319-10602-1_48}). The $F$-score~(\cite{10.3115/1072017.1072026}), also known as dice coefficient, is a widely used metric for ranking purposes, and consists of the harmonic mean of precision $P$ and recall $R$ metrics
\begin{equation}
   F = \frac{2}{1/P + 1/R}
\end{equation}
such that 
\begin{equation}
   P = \frac{tp}{tp + fp} \qquad \qquad R = \frac{tp}{tp + fn} 
\end{equation}
where $tp$ (true positives) is the number of correctly classified fire pixels, $fp$ (false positives) the number of non-fire pixels incorrectly classified as fire, and $fn$ (false negatives) the number of fire pixels incorrectly classified as non-fire.

We also report the Intersection-Over-Union (IoU) metric~(\cite{NIPS2015_4e4e53aa}), also known as Jaccard index

\begin{equation}
  \textrm{IoU} = \frac{tp}{tp + fn + fp} 
\end{equation}

The IoU metric conveys the same information as the $F$-score --- $\textrm{IoU} = F / (2 - F)$) --- but it is also widely in many segmentation challenges, so for clarity we report both values. All metrics were computed as per the COCO evaluation implementation: $tp$, $fp$ and $fn$ values are accumulated for all the images and the metrics are computed only once for the entire dataset (i.e.~the reported values are not the average per-patch performance, but the global per-pixel performance).

As in the KITTI challenge~(\cite{8702174}), we ranked the algorithm's performances according to the $F$-score metric. 

One common metric that we do not list in our experiments is the pixel accuracy metric, which takes into account the number of true negatives (non-fire pixels detected as such). In our tests, we observed a large class imbalance between fire and non-fire pixels --- more than 99\% of the total samples are non-fire pixels --- and that always resulted in very high accuracy --- even for an extreme case of a detector that fails to detect any fire pixels could still seem to perform reasonably well, making pixel accuracy a misleading metric.


\subsection{Performance on the automatically segmented patches}
\label{section.experiments1}

For the first round of experiments, we used the CNN architectures trained and validated with 50\% of the image patches against the other 50\% reserved for testing, in order to show how well each trained architecture can approximate the original segmentations obtained by the~\cite{SCHROEDER2016210}, \cite{MURPHY201678} and \cite{doi:10.1080/17538947.2017.1391341} conditions, as well as combinations of their outputs (their intersection and a ``best-of-three'' voting). In total, we had 15 testing scenarios, as shown in Table~\ref{tab:performance}, with 3 CNN architectures (U-Net (10c), U-Net (3c) and U-Net-Light (3c)) to approximate the 5 target segmentation masks.

As can be seen, all architectures were able to reproduce the behavior of the different algorithms reasonably well. It is interesting to note that a larger network (U-Net (10c)) is not necessarily superior to a network that uses a reduced number of channels (U-Net (3c)), nor to a less complex network with a reduced number of channels (U-Net-Light (3c)). For most scenarios the performances were similar, with some small variations that may be attributed to the initial conditions of the trained networks, sampling, and other random factors. That means it is likely that this particular application does not demand very complex architectures, and more importantly, that channels $c_7$, $c_6$ and $c_2$ indeed contain most of the information needed for detecting active fires. Moreover, there is a tendency that models with higher precision have lower recall, and vice-versa, but not to a large margin in most cases, which shows the learning procedure is balancing false and missed detections.

 \begin{table}[!htb] \footnotesize
   \setlength{\tabcolsep}{6pt}
   \renewcommand{\arraystretch}{1.1}
   \centering
   \caption{Active fire recognition performance: we used the set of conditions from \cite{SCHROEDER2016210}, \cite{MURPHY201678} and \cite{doi:10.1080/17538947.2017.1391341}  to train three convolutional neural networks for active fire segmentation: U-Net (10c) uses as input a 10-channel image from Landsat-8; U-Net (3c) uses as input a 3-channel image, containing only bands $c_7, c_6$, and $c_2$, which have a good response for active fire; and U-Net-Light (3c) is a reduced version of the 3-channel U-Net. The performances are related to the CNN ability to reproduce the masks from \cite{SCHROEDER2016210}, \cite{MURPHY201678}, \cite{doi:10.1080/17538947.2017.1391341}, and also the intersection and consensus of their masks, in a testing set never seen by these architectures.}
   \label{tab:performance}
   \begin{tabular}{cccccc}
    \hline \hline
     Mask  &      CNN Architecture          & \multicolumn{4}{c}{Metrics (\%)} \\ \cline{3-6}
   &       & $\;\;\;P\;\;\;$ & $\;\;\;R\;\;\;$ &  \;IoU\;\; & \;\;$F$\;\; \\
   \hline \hline
     & U-Net (10c) & 86.8 & \textbf{89.7} &  78.9 & 88.2 \\
 Schroeder~\etal  & U-Net (3c) & 89.8 & 88.8 &  \textbf{80.7} &  \textbf{89.3} \\
    &  U-Net-Light (3c) & \textbf{90.8} & 86.1 & 79.2 & 88.4 \\
   \hline
    & U-Net (10c) & \textbf{93.6} & 92.5 &  87.0 & 93.0 \\
 Murphy~\etal & U-Net (3c) & 89.1 & \textbf{97.6} & 87.2 & 93.2 \\
    & U-Net-Light (3c) & 92.6 & 95.1 & \textbf{88.4} & \textbf{93.8} \\
   \hline
    & U-Net (10c) & \textbf{84.6} & \textbf{94.1} &  \textbf{80.3} & \textbf{89.1}\\
 Kumar-Roy  & U-Net (3c) & 84.2 & 90.6 &  77.5 & 87.3 \\
    & U-Net-Light (3c) & 76.8 & 93.2 &  72.7 & 84.2 \\
   \hline \\[-3.5mm]
  \multirow{3}{*}{\textcolor{white}{.}Intersection\textcolor{white}{.}$\left\{\begin{array}{l}
      \hspace{-6pt}\textrm{Schroeder~\etal} \\
      \hspace{-6pt}\textrm{Murphy~\etal} \\
      \hspace{-6pt}\textrm{Kumar-Roy}
    \end{array}\right.$}  & U-Net (10c) & 84.4 & \textbf{99.7} &  84.2 & 91.4 \\
  &  U-Net (3c) & \textbf{93.4} & 92.4 & \textbf{86.7} & \textbf{92.9} \\
  &  U-Net-Light (3c) & 87.4 & 97.3 & 85.4 & 92.1  \\[+0.8mm]
  \hline \\[-3.5mm]
  \multirow{3}{*}{\textcolor{white}{...}Voting\textcolor{white}{.....} $\left\{
    \begin{array}{l}
     \hspace{-6pt}\textrm{Schroeder~\etal} \\
     \hspace{-6pt}\textrm{Murphy~\etal} \\
     \hspace{-6pt}\textrm{Kumar-Roy}
    \end{array}\right.$}  & U-Net (10c) & \textbf{92.9} & 95.5 &  \textbf{89.0} & \textbf{94.2} \\
  &  U-Net (3c)  & 91.9 & 95.3 &  87.9 & 93.6 \\
  &  U-Net-Light (3c) & 90.2 & \textbf{96.5} & 87.3 & 93.2 \\[+0.8mm] \hline \hline
   \end{tabular}
\end{table}

\begin{figure}[!htb]
   \setlength{\tabcolsep}{2pt}
   \renewcommand{\arraystretch}{1.2}
   \centering
   \begin{tabular}{ccccc}
   {\footnotesize Landsat-8} & {\footnotesize Schroeder~\etal} &  {\footnotesize U-Net (10c)} & {\footnotesize U-Net (3c)} & {\footnotesize U-Net-Light (3c)}\\
   \includegraphics[width=0.190\textwidth]{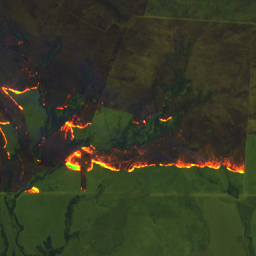} &
   \includegraphics[width=0.190\textwidth]{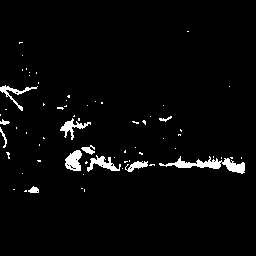}  &
   \includegraphics[width=0.190\textwidth]{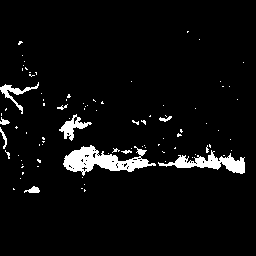} &
   \includegraphics[width=0.190\textwidth]{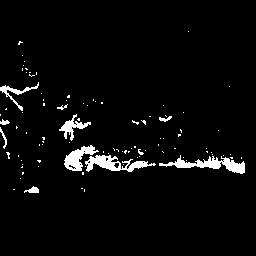} &
   \includegraphics[width=0.190\textwidth]{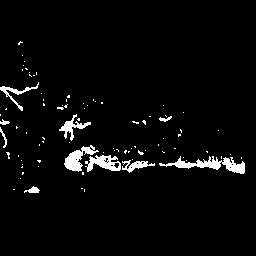} \\
   \includegraphics[width=0.190\textwidth]{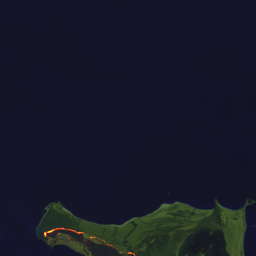} &
   \includegraphics[width=0.190\textwidth]{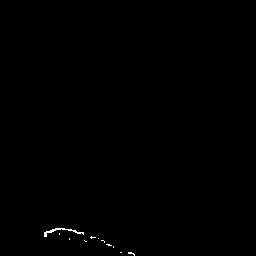}  &
   \includegraphics[width=0.190\textwidth]{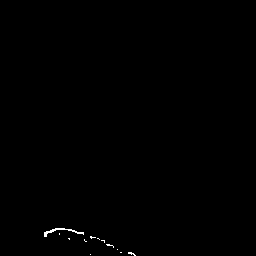} &
   \includegraphics[width=0.190\textwidth]{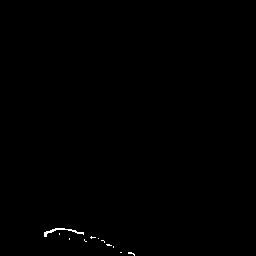} &
   \includegraphics[width=0.190\textwidth]{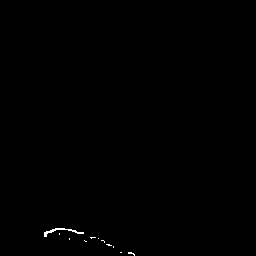}
   \end{tabular}
 \caption{Active fire segmentation results obtained by the algorithm from \cite {SCHROEDER2016210}, and by the networks trained using those outputs as the target masks.}
 \label{fig:cnn_schroeder}
\end{figure}

\begin{figure}[!htb]
   \setlength{\tabcolsep}{2pt}
   \renewcommand{\arraystretch}{1.2}
   \centering
   \begin{tabular}{ccccc}
   {\footnotesize Landsat-8} & {\footnotesize Murphy~\etal} &  {\footnotesize U-Net (10c)} & {\footnotesize U-Net (3c)} & {\footnotesize U-Net-Light (3c)}\\
   \includegraphics[width=0.190\textwidth]{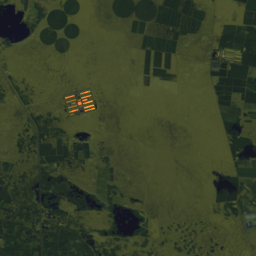} &
   \includegraphics[width=0.190\textwidth]{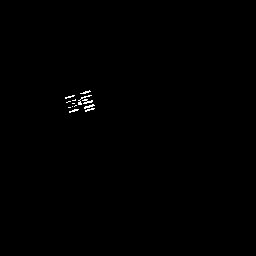}  &
   \includegraphics[width=0.190\textwidth]{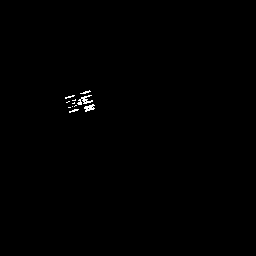} &
   \includegraphics[width=0.190\textwidth]{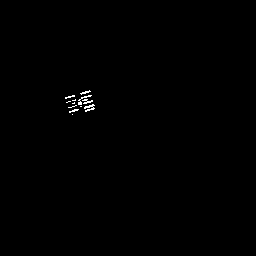} &
   \includegraphics[width=0.190\textwidth]{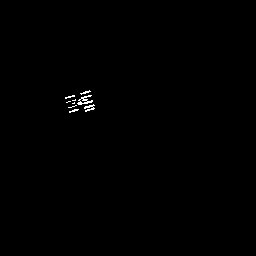} \\
   \includegraphics[width=0.190\textwidth]{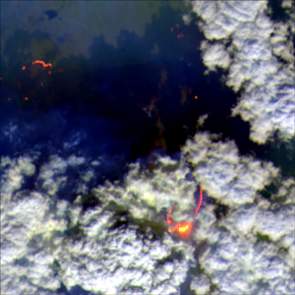} &
   \includegraphics[width=0.190\textwidth]{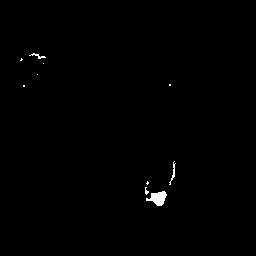}  &
   \includegraphics[width=0.190\textwidth]{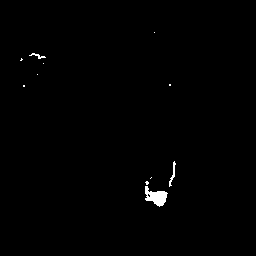} &
   \includegraphics[width=0.190\textwidth]{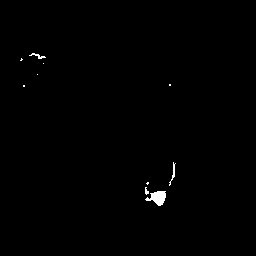} &
   \includegraphics[width=0.190\textwidth]{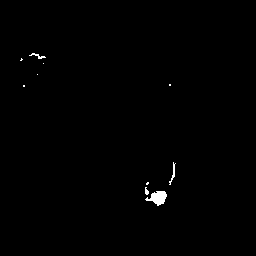}
   \end{tabular}
 \caption{Active fire segmentation results obtained by the algorithm from \cite {MURPHY201678}, and by the networks trained using those outputs as the target masks.}
 \label{fig:cnn_murphy}
\end{figure}

\begin{figure}[!htb]
   \setlength{\tabcolsep}{2pt}
   \renewcommand{\arraystretch}{1.2}
   \centering
   \begin{tabular}{ccccc}
   {\footnotesize Landsat-8} & {\footnotesize Kumar-Roy} &  {\footnotesize U-Net (10c)} & {\footnotesize U-Net (3c)} & {\footnotesize U-Net-Light (3c)}\\
   \includegraphics[width=0.190\textwidth]{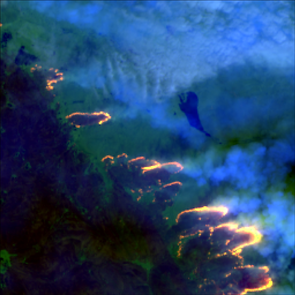} &
   \includegraphics[width=0.190\textwidth]{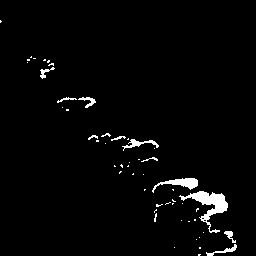}  &
   \includegraphics[width=0.190\textwidth]{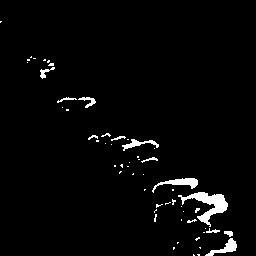} &
   \includegraphics[width=0.190\textwidth]{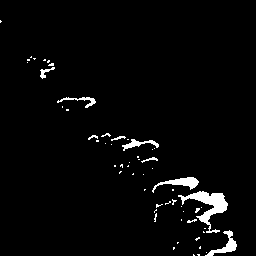} &
   \includegraphics[width=0.190\textwidth]{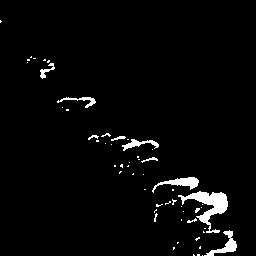} \\
   \includegraphics[width=0.190\textwidth]{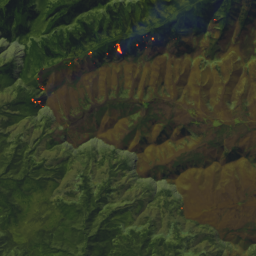} &
   \includegraphics[width=0.190\textwidth]{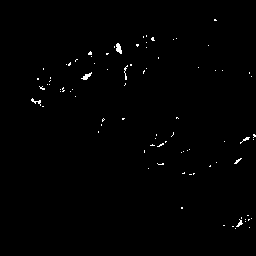}  &
   \includegraphics[width=0.190\textwidth]{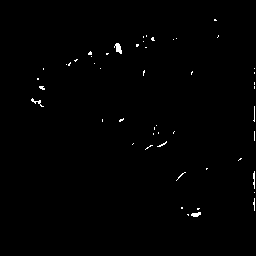} &
   \includegraphics[width=0.190\textwidth]{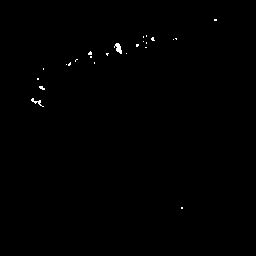} &
   \includegraphics[width=0.190\textwidth]{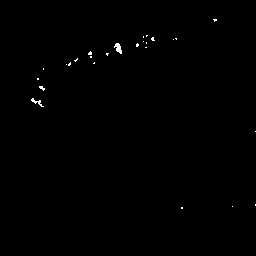}
   \end{tabular}
 \caption{Active fire segmentation results obtained by the algorithm from \cite {doi:10.1080/17538947.2017.1391341}, and by the networks trained using those outputs as the target masks.}
 \label{fig:cnn_kumar_roy}
\end{figure}

Figures \ref {fig:cnn_schroeder}, \ref {fig:cnn_murphy} and \ref {fig:cnn_kumar_roy} compare the outputs produced by the CNNs to the outputs of, respectively, the~\cite {SCHROEDER2016210},~\cite {MURPHY201678} and~\cite {doi:10.1080/17538947.2017.1391341} sets of conditions, for image samples.  It can be seen that the CNNs indeed succeed in approximating the target masks, especially considering the presence of fire in a certain area, with most differences at the pixel level being small. Even the 3-channel architectures are successful in most cases, once again confirming that channels $c_7$, $c_6$ and $c_2$ contain most of the information needed for detecting active fires. 

The second row from Figure~\ref {fig:cnn_kumar_roy}, which has several small clusters of fire pixels detected by the original algorithm and the 10-channel CNN, but which were not found by the 3-channel networks (and which are, consistently, hard to see in the 3-channel visualization), is an exception  of good approximation for the networks with 3-channels.

\begin{figure}[!htb]
   \setlength{\tabcolsep}{2pt}
   \renewcommand{\arraystretch}{0.7}
   \centering
   \begin{tabular}{ccc}
   {\scriptsize Landsat-8} & {\scriptsize Schroeder (S)} &  {\scriptsize Murphy (M)}  \\
   \includegraphics[width=0.230\textwidth]{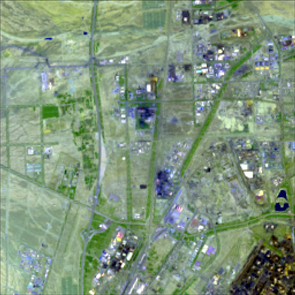} &
   \includegraphics[width=0.230\textwidth]{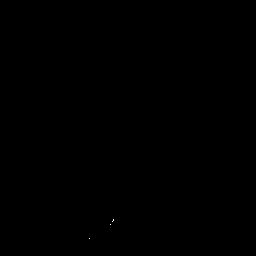}  &
   \includegraphics[width=0.230\textwidth]{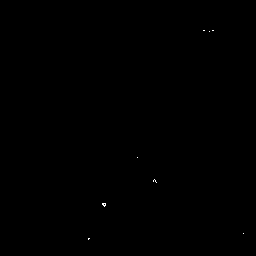} \\
   {\scriptsize Kumar-Roy (K-R)} & {\scriptsize Intersection (S, M, K-R)} & {\scriptsize U-Net (3c)} \\
   \includegraphics[width=0.230\textwidth]{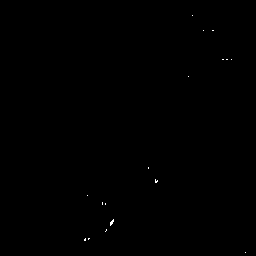} &
   \includegraphics[width=0.230\textwidth]{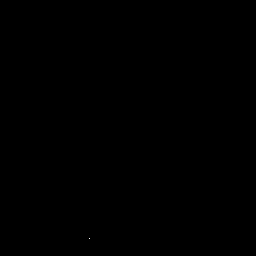}  &
   \includegraphics[width=0.230\textwidth]{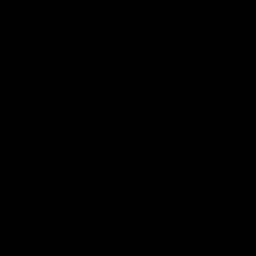} 
   \end{tabular}
 \caption{The~\cite{SCHROEDER2016210}, \cite{MURPHY201678} and \cite{doi:10.1080/17538947.2017.1391341} algorithms had false detections inside a city, but their intersection and the 3-channel U-Net trained to approximate it were able to avoid them.}
 \label{fig:cnn_intersection}
\end{figure}

\begin{figure}[!htb]
   \setlength{\tabcolsep}{2pt}
   \renewcommand{\arraystretch}{0.7}
   \centering
   \begin{tabular}{ccc}
   {\scriptsize Landsat-8} & {\scriptsize Schroeder (S)} &  {\scriptsize Murphy (M)}  \\
   \includegraphics[width=0.230\textwidth]{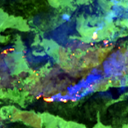} &
   \includegraphics[width=0.230\textwidth]{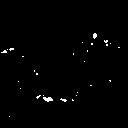}  &
   \includegraphics[width=0.230\textwidth]{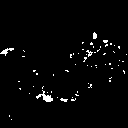} \\
   {\scriptsize Kumar-Roy (K-R)} & {\scriptsize Voting (S, M, K-R)} & {\scriptsize U-Net (3c)} \\
   \includegraphics[width=0.230\textwidth]{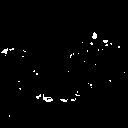} &
   \includegraphics[width=0.230\textwidth]{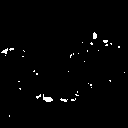}  &
   \includegraphics[width=0.230\textwidth]{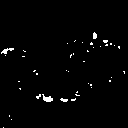} 
   \end{tabular}
 \caption{The 3-channel U-Net was able to approximate a ``best-of-three'' voting scheme that takes the outputs from the~\cite{SCHROEDER2016210}, \cite{MURPHY201678} and \cite{doi:10.1080/17538947.2017.1391341} algorithms. The voting scheme reduces errors caused by artifacts produced by only one of the algorithms.}
 \label{fig:cnn_voting}
\end{figure}

In Figures~\ref {fig:cnn_intersection} and~\ref {fig:cnn_voting}, we compare the outputs from the three sets of conditions to their combination (first their intersection, then the ``best-of-three'' voting scheme), as well as the output of the 3-channel U-Net trained to approximate that combination. 


Figure~\ref {fig:cnn_intersection} shows that, while the three sets of conditions may incorrectly detect active fires inside urban settlements, their intersection is usually more robust (although that may also result in lower recall). As for Figure~\ref {fig:cnn_voting}, it shows the voting scheme helps reducing behaviors that are associated with only one of the sets of conditions, such as the tendency of the Murphy et al. contitions to produce thicker fire clusters and, in this example, the ``hole'' in the lower fire region from the Kumar-Roy conditions. In both cases, the trained CNNs were successful in approximating the combined masks. Another question brought by these images is: how do the 3 sets of conditions compare to each other, and to the networks that learn to combine them? This matter will be analyzed in more depth in Section~\ref {section.experiments_manual}.


\subsection{Performance on the manually annotated images}
\label {section.experiments_manual}

For the second round of experiments, we fed the CNN models trained in the first round (Section~\ref{section.experiments1}) with Landsat-8 image patches from the manually annotated dataset, and compared the results with the hand-segmented masks. For reference, the outputs from the~\cite{SCHROEDER2016210}, \cite{MURPHY201678} and \cite{doi:10.1080/17538947.2017.1391341} algorithms were also compared to the hand-segmented masks. This allows us to observe how well the networks learned from the automatically segmented patches perform when compared to a human specialist (one must keep in mind, however, that different specialists would produce distinct segmentations, since it is very subjective sometimes to separate the frontiers of burnt areas from areas that are still burning, all this added to the presence of smoke and many levels of fire intensity). For works on burnt areas the reader can refer to \cite{CHEN202063,MALAMBO2020107}.

\begin{figure}[!htb]
  \hspace{-10pt}
  \begin{tikzpicture}[]

   \node[inner sep=0pt] (img) at (2.14,14.43) { \includegraphics[width=2.6cm,height=2.6cm]{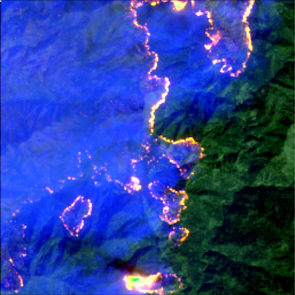}};

   \node[inner sep=0pt] (img) at (5.66,14.43) { \includegraphics[width=2.6cm,height=2.6cm]{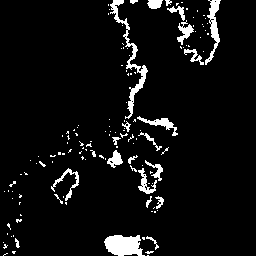}};
   
   \draw[fill=blue!10!white,blue!10!white] (-2.93,0.0) rectangle (-0.24,12.5);
   \draw[fill=gray!7,gray!7] (-0.24,0.0) rectangle (2.52,12.5);
   \draw[fill=blue!10!white,blue!10!white] (2.52,0.0) rectangle (5.28,12.5);
   \draw[fill=gray!7,gray!7] (5.28,0.0) rectangle (8.04,12.5);
   \draw[fill=blue!10!white,blue!10!white] (8.04,0.0) rectangle (10.73,12.5);


   \node[rotate=+90] at (-3.12,1.45) {\scriptsize U-Net-Light (3c)};
   
   \node[inner sep=0pt] (img) at (-1.62,1.45) { \includegraphics[width=2.6cm,height=2.6cm]{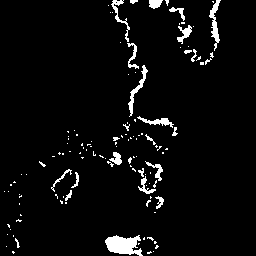}};
   
   \node[inner sep=0pt] (img) at (+1.14,1.45) { \includegraphics[width=2.6cm,height=2.6cm]{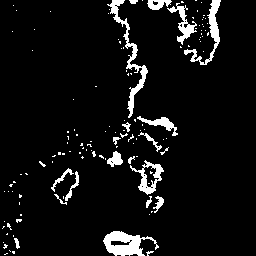}};
   
   \node[inner sep=0pt] (img) at (+3.90,1.45) { \includegraphics[width=2.6cm,height=2.6cm]{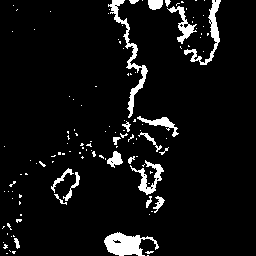}};
   
   \node[inner sep=0pt] (img) at (+6.66,1.45) { \includegraphics[width=2.6cm,height=2.6cm]{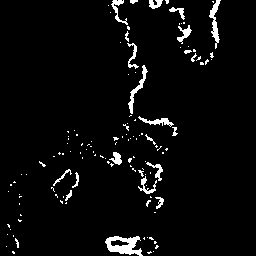}};
   
   \node[inner sep=0pt] (img) at (+9.42,1.45) { \includegraphics[width=2.6cm,height=2.6cm]{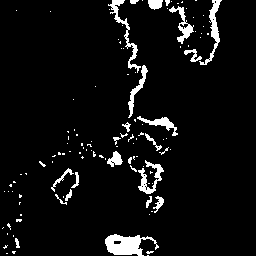}};
   
   
   \node[rotate=+90] at (-3.12,4.4) {\scriptsize U-Net (3c)};

   \node[inner sep=0pt] (img) at (-1.62,4.4) { \includegraphics[width=2.6cm,height=2.6cm]{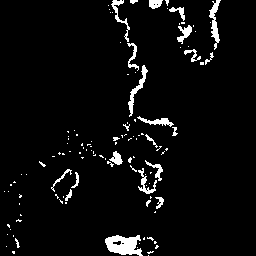}};
   
   \node[inner sep=0pt] (img) at (+1.14,4.4) { \includegraphics[width=2.6cm,height=2.6cm]{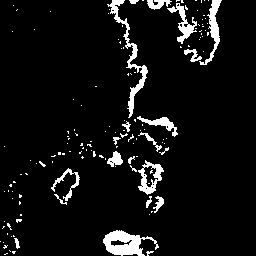}};
   
   \node[inner sep=0pt] (img) at (+3.90,4.4) { \includegraphics[width=2.6cm,height=2.6cm]{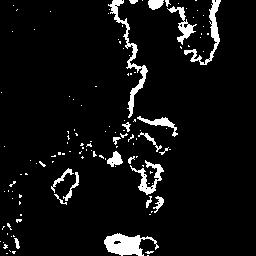}};
   
   \node[inner sep=0pt] (img) at (+6.66,4.4) { \includegraphics[width=2.6cm,height=2.6cm]{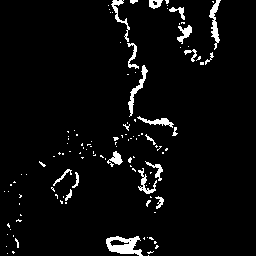}};
   
   \node[inner sep=0pt] (img) at (+9.42,4.4) { \includegraphics[width=2.6cm,height=2.6cm]{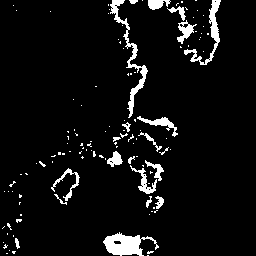}};
   
   
   \node[rotate=+90] at (-3.12,7.35) {\scriptsize U-Net (10c)};
   
   \node[inner sep=0pt] (img) at (-1.62,7.35) { \includegraphics[width=2.6cm,height=2.6cm]{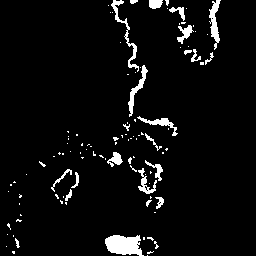}};
   
   \node[inner sep=0pt] (img) at (+1.14,7.35) { \includegraphics[width=2.6cm,height=2.6cm]{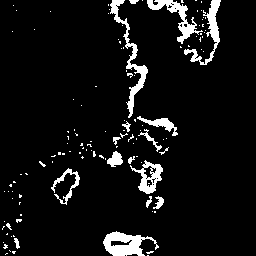}};
   
   \node[inner sep=0pt] (img) at (+3.90,7.35) { \includegraphics[width=2.6cm,height=2.6cm]{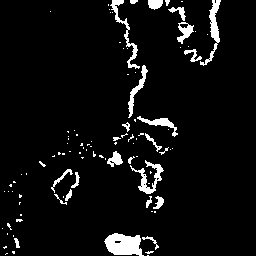}};
   
   \node[inner sep=0pt] (img) at (+6.66,7.35) { \includegraphics[width=2.6cm,height=2.6cm]{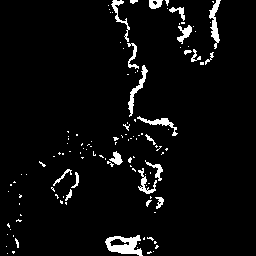}};
   
   \node[inner sep=0pt] (img) at (+9.42,7.35) { \includegraphics[width=2.6cm,height=2.6cm]{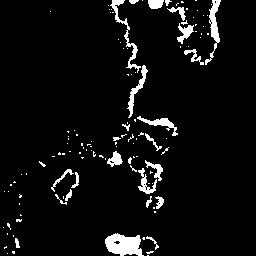}};
   
   
   \node[rotate=+90] at (-3.12,10.3) {\scriptsize Original Conditions};

   \node[inner sep=0pt] (img) at (-1.62,10.3) { \includegraphics[width=2.6cm,height=2.6cm]{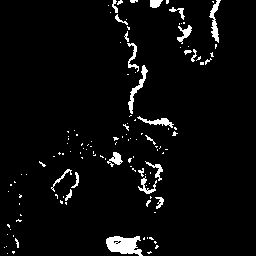}};
   
   \node[inner sep=0pt] (img) at (+1.14,10.3) { \includegraphics[width=2.6cm,height=2.6cm]{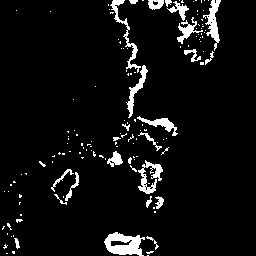}};
   
   \node[inner sep=0pt] (img) at (+3.90,10.3) { \includegraphics[width=2.6cm,height=2.6cm]{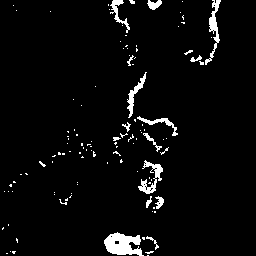}};
   
   \node[inner sep=0pt] (img) at (+6.66,10.3) { \includegraphics[width=2.6cm,height=2.6cm]{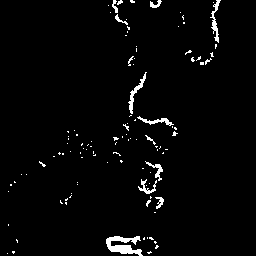}};
   
   \node[inner sep=0pt] (img) at (+9.42,10.3) { \includegraphics[width=2.6cm,height=2.6cm]{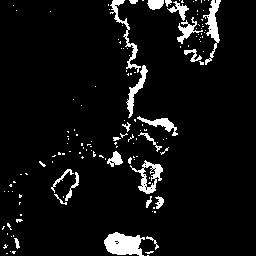}};
   
   
   \node[rotate=0] at (-1.62,12.0) {\scriptsize Schroeder~\etal};
   
   \node[rotate=0] at (+1.14,12.0) {\scriptsize Murphy~\etal};
   
   \node[rotate=0] at (+3.90,12.0) {\scriptsize Kumar-Roy};
   
   \node[rotate=0] at (+6.66,12.0) {\scriptsize Intersection};
   
   \node[rotate=0] at (+9.42,12.0) {\scriptsize Voting};
   
   \node[rotate=0] at (+2.14,15.9) {\scriptsize Landsat-8 image patch};
   
   \node[rotate=0] at (+5.66,15.9) {\scriptsize Manual annotations};
   
   \end{tikzpicture}
\caption{Active fire segmentation results produced by the~\cite{SCHROEDER2016210},~\cite{MURPHY201678} and~\cite{doi:10.1080/17538947.2017.1391341} sets of conditions, the deep networks trained to approximate them, and their combined outputs (intersection and ``best of three'' voting), for a Landsat-8 image patch.}
\label{fig:manual_result}
\end{figure}

Table~\ref {tab:performance_manual} shows the performances obtained for each one of the 15 scenarios involving CNN architectures, as described in section~\ref{section.experiments1}, as well as, the performances for the original set of conditions of~\cite{SCHROEDER2016210}, \cite{MURPHY201678} and \cite{doi:10.1080/17538947.2017.1391341} without the use of any CNN architecture. 
As can be seen, the best F-scores were around 90\% --- this is expected, as there are biases from both the human specialist labeling the data and the algorithms themselves, which are tuned to perform under certain assumptions. 
As an example, we also show in Figure~\ref {fig:manual_result} a selected image patch from Landsat-8, with a large area of active fire, along with its associated manual segmentation, and also the segmentation masks produced for each scenario being considered.

\begin{table}[!htb] \footnotesize
   \setlength{\tabcolsep}{6pt}
   \renewcommand{\arraystretch}{1.1}
   \centering
   \caption{Active fire recognition performance against \textbf{manual annotations} for 13 Landsat-8 scenes. The entries with --- do not use a CNN, the performances are computed directly from the \cite{SCHROEDER2016210}, \cite{MURPHY201678} and \cite{doi:10.1080/17538947.2017.1391341} sets of conditions against the manual annotations. The remaining entries correspond to deep networks trained with masks automatically extracted using these sets of conditions. All entries were compared against the same manual annotations and the boldface values show the best performance for each metric.}
   \label{tab:performance_manual}
   \begin{tabular}{cccccc}
   \hline \hline
     Mask  &      CNN Architecture          & \multicolumn{4}{c}{Metrics (\%)} \\ \cline{3-6}
   &       & $\;\;\;P\;\;\;$ & $\;\;\;R\;\;\;$ & \;IoU\;\; & \;\;$F$\;\; \\
   \hline \hline 
     Schroeder~\etal  &  --- & 88.1  & 70.2  & 64.1 & 78.1\\ \hdashline[0.6pt/2pt]
     & U-Net (10c) & 86.9  & 88.4  &  78.0 &   87.7  \\
 Schroeder~\etal  & U-Net (3c) & 89.5  & 80.5  & 73.6 & 84.8 \\
    &  U-Net-Light (3c) & 89.0 & 78.8  & 71.8  & 83.6    \\
   \hline 
  Murphy~\etal  & ---  & 76.6  &  96.1 & 74.3  & 85.2  \\ \hdashline[0.6pt/2pt] 
    & U-Net (10c) & 74.8 & 96.9 & 73.0 & 84.4\\ 
 Murphy~\etal & U-Net (3c) & 72.7 & \textbf{97.2} & 71.2 & 83.2 \\
    & U-Net-Light (3c) & 75.5 & 96.9 & 73.7 & 84.9 \\
   \hline 
 Kumar-Roy  &  --- & 82.3  &  68.4 &  59.7  & 74.7 \\ \hdashline[0.6pt/2pt]  
    & U-Net (10c) & 82.2 & 94.2 & 78.3 & 87.8 \\
 Kumar-Roy  & U-Net (3c) & 84.5 & 93.4 & 79.8 &  88.8 \\
    & U-Net-Light (3c) &  78.8 & 96.9 & 76.9 & 86.9 \\
   \hline  \\[-3.5mm]
  \multirow{3}{*}{\textcolor{white}{.}Intersection\textcolor{white}{.}$\left\{\begin{array}{l}
      \hspace{-6pt}\textrm{Schroeder~\etal} \\
      \hspace{-6pt}\textrm{Murphy~\etal} \\
      \hspace{-6pt}\textrm{Kumar-Roy}
\end{array}\right.$} & U-Net (10c) & \textbf{91.8} & 75.4 & 70.6 & 82.8 \\
  &  U-Net (3c) & 90.8  & 73.5  & 68.4  &  81.2  \\
  &  U-Net-Light (3c) & 90.8  & 72.8  & 67.8 & 80.8  \\[+0.8mm]
  \hline \\[-3.5mm]
  \multirow{3}{*}{\textcolor{white}{...}Voting\textcolor{white}{.....} $\left\{
    \begin{array}{l}
     \hspace{-6pt}\textrm{Schroeder~\etal} \\
     \hspace{-6pt}\textrm{Murphy~\etal} \\
     \hspace{-6pt}\textrm{Kumar-Roy}
\end{array}\right.$}  & U-Net (10c) & 83.6 & 94.0 & 79.3 & 88.5 \\
  &  U-Net (3c) & 86.4  & 93.0 & 81.1 & 89.6 \\
  &  U-Net-Light (3c) & 87.2 & 92.4 & \textbf{81.4} & \textbf{89.7} \\ \hline \hline
   \end{tabular}
\end{table}

As can be seen in Table~\ref {tab:performance_manual}, the original conditions from~\cite{MURPHY201678} performed better than those from~\cite{SCHROEDER2016210} and~\cite{doi:10.1080/17538947.2017.1391341} in the manually annotated dataset, mostly due to its tendency to detect more fire pixels than the other conditions, resulting in a very high recall, albeit at the cost of a lower precision (i.e.~it is more prone to false detections). The~\cite{doi:10.1080/17538947.2017.1391341} original conditions had lower precision and recall compared to the~\cite{SCHROEDER2016210} conditions --- in fact, it fails to detect several fire pixels in the example of Figure~\ref {fig:manual_result}, an issue that will be further discussed briefly. All the sets of conditions had problems detecting the core of the fire at the bottom of the image, which has very high intensity.

As for the CNNs, most differences between them and the original sets of conditions were actually positive, with the deep networks performing better than the~\cite{SCHROEDER2016210} and \cite{doi:10.1080/17538947.2017.1391341} sets of conditions, with higher recall but without a corresponding hit on precision, as shown in Table~\ref {tab:performance_manual}. In Figure~\ref {fig:manual_result}, the networks trained to approximate the~\cite{doi:10.1080/17538947.2017.1391341} conditions were able to detect many fire pixels missed by the original algorithm. The deep networks have also shown a higher recall (but lower precision) compared to the~\cite{MURPHY201678} conditions. As expected, by taking the intersection of three network outputs, we obtained the highest precision but a very lower recall, since this combination keeps only those pixels that all three networks agree are fire pixels, that is, it is very restrictive. The best overall performance --- including the original sets of conditions --- was obtained by the voting scheme applied to the outputs of three networks, which resulted in high recall without a great impact on the precision, with the 3-channel light U-Net being the one whose results were the most similar to the manual segmentations, but with no significant difference for the larger 3-channel U-Net. We emphasize that these networks were trained using only automatically segmented samples, so the improved performance cannot be caused by some bias learned from the manually annotated patches, which were only seen in the tests.

\begin{figure}[!htb]
   \setlength{\tabcolsep}{2pt}
   \renewcommand{\arraystretch}{1.2}
   \centering
   \label{tab:results_figure4}
   \begin{tabular}{ccccc}
   {\scriptsize Landsat-8 image} & {\scriptsize Schroeder et al.} &  {\scriptsize U-Net (10c)} & {\scriptsize U-Net (3c)} & {\scriptsize U-Net-Light (3c)}\\
   \includegraphics[width=0.190\textwidth]{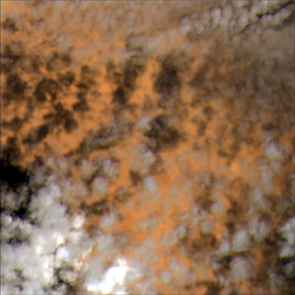} &
   \includegraphics[width=0.190\textwidth]{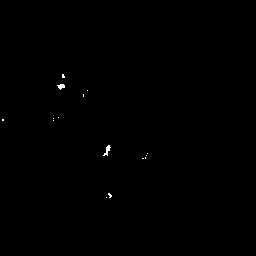} &
   \includegraphics[width=0.190\textwidth]{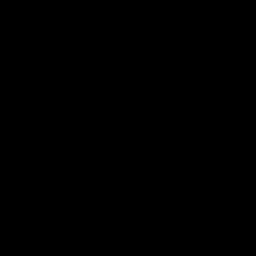} &
   \includegraphics[width=0.190\textwidth]{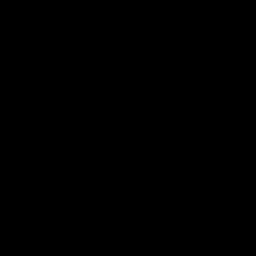} &
   \includegraphics[width=0.190\textwidth]{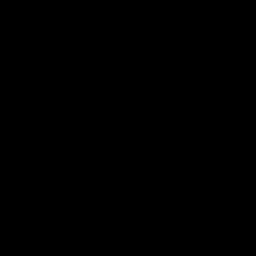} \\
   \multicolumn{5}{c}{\scriptsize (a) Segmentations from Schoeder et al. conditions and for U-Net architectures based of his conditions.}\\
   {\scriptsize Landsat-8 image} & {\scriptsize Murphy et al.} &  {\scriptsize U-Net (10c)} & {\scriptsize U-Net (3c)} & {\scriptsize U-Net-Light (3c)}\\
   \includegraphics[width=0.190\textwidth]{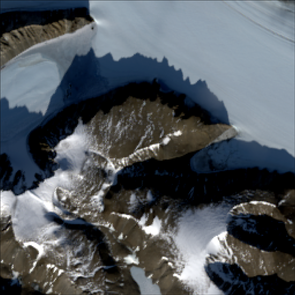} &
   \includegraphics[width=0.190\textwidth]{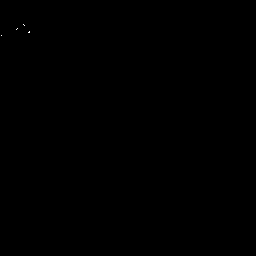} &
   \includegraphics[width=0.190\textwidth]{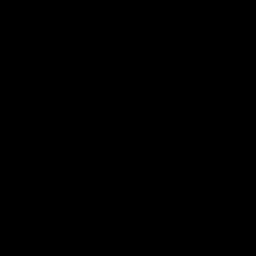} &
   \includegraphics[width=0.190\textwidth]{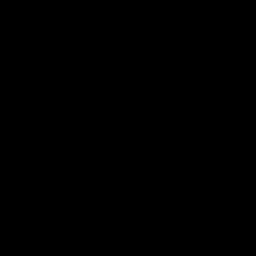} &
   \includegraphics[width=0.190\textwidth]{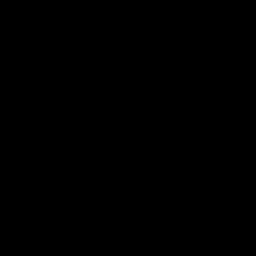} \\
   \multicolumn{5}{c}{\scriptsize (b) Segmentations from Murphy conditions and for U-Net architectures based of his conditions.}\\
   {\scriptsize Landsat-8 image} & {\scriptsize Kumar-Roy} &  {\scriptsize U-Net (10c)} & {\scriptsize U-Net (3c)} & {\scriptsize U-Net-Light (3c)}\\
   \includegraphics[width=0.190\textwidth]{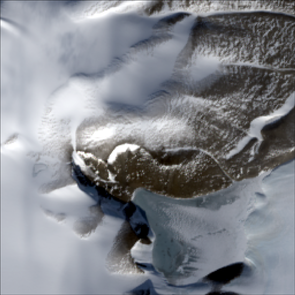} &
   \includegraphics[width=0.190\textwidth]{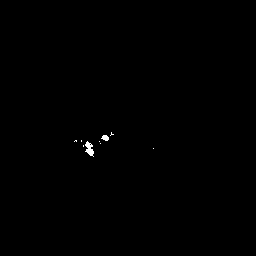} &
   \includegraphics[width=0.190\textwidth]{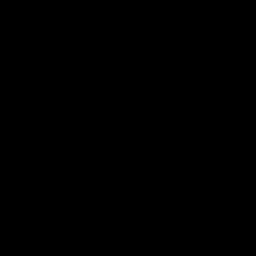} &
   \includegraphics[width=0.190\textwidth]{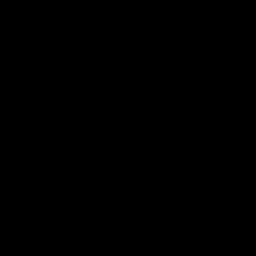} &
   \includegraphics[width=0.190\textwidth]{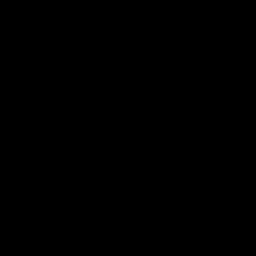} \\
   \multicolumn{5}{c}{\scriptsize (c) Segmentations from Kumar-Roy conditions and for U-Net architectures based of his conditions.}\\
   {\scriptsize Landsat-8 image} & {\scriptsize Schroeder et al.} &  {\scriptsize Murphy et al.} & {\scriptsize Kumar-Roy} & {\scriptsize U-Net (10c)}\\
   \includegraphics[width=0.190\textwidth]{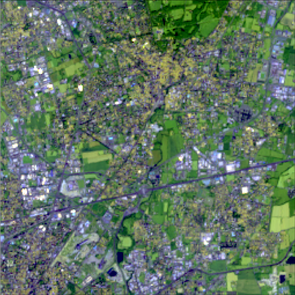} &
   \includegraphics[width=0.190\textwidth]{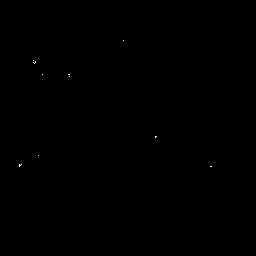} &
   \includegraphics[width=0.190\textwidth]{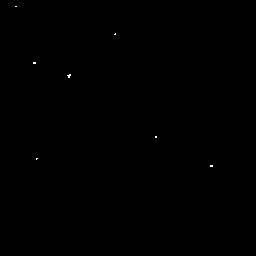} &
   \includegraphics[width=0.190\textwidth]{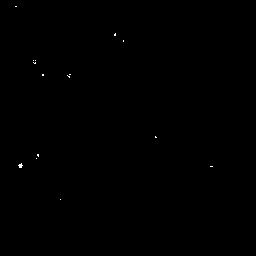} &
   \includegraphics[width=0.190\textwidth]{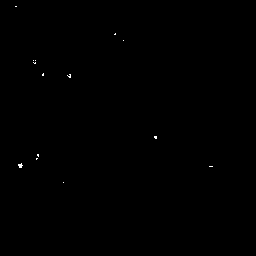} \\
   \multicolumn{5}{c}{\scriptsize (d) Segmentations from Schoeder et al., Murphy et al, Kumar-Roy and a U-Net architecture.}\\
   \end{tabular}
 \caption{Classification errors: (a) region from Sahara desert WRS: 189/046, date 2020/09/18; (b) and (c) from Greenland, WRS: 003/006, date: 2020/09/11; and (d) city of Milan, Italy, WRS: 193/029, date: 2020/09/14, where all original conditions and CNN architectures reported misclassifications for
bright and/or reflective urban elements.}
 \label{fig:classification_errors}
\end{figure}

Besides the quantitative results discussed so far, we analyzed a number of individual cases where the performance or recall obtained by one of the approaches was particularly low. One example is the segmentation from the~\cite{doi:10.1080/17538947.2017.1391341} original conditions shown in Figure~\ref {fig:manual_result}, which missed many fire pixels. We observed that these pixels were in fact detected as fire pixels, but were later discarded after being regarded as water pixels, due to some very similar values in bands $\rho_2$ to $\rho_5$. Other cases are presented in Figure~\ref {fig:classification_errors}, which shows failure cases for each one of the sets of conditions. In the first row, Figure~\ref {fig:classification_errors}(a), the~\cite{SCHROEDER2016210} conditions produced false detections in a Sahara desert region. In the second and third rows, Figure~\ref {fig:classification_errors}(b, c), the~\cite{MURPHY201678} and \cite{doi:10.1080/17538947.2017.1391341} conditions produced false detections in some Greenland regions. In the map previously shown in Figure~\ref{fig:heatmap} these errors can be seen as green dots in the Sahara region and blue and magenta dots in Greenland --- but note the map refers to data from August 2020, while the examples in Figure~\ref {fig:classification_errors} were taken in September 2020, indicating that the algorithms are consistently producing false detections in some situations. 

By analyzing these cases in further detail, by looking at each step and component from the conditions, we noticed that most errors were not caused by fundamental problems with the conditions themselves, but by reflectance values that appear very close to the thresholds from one rule or another. For example, some errors could be avoided if the $0.2$ threshold from one unambiguous fire condition from~\cite {SCHROEDER2016210} was changed to $0.194$ --- but that could potentially also lead to real fires not being detected.  This highlights one of the challenges faced when handcrafting sets of rules: some situations must be simplified to keep the algorithm readable and understandable, so thresholds are usually rounded to 1 or 2 decimal places, and only the most important relations between bands are considered. Machine learning techniques are able to encode more complex rules, with more precise weights, coefficients and thresholds; although this may also mean that the learned rules and relations can be hard to understand and describe by a human. In any case, the deep networks were able to avoid the errors in the samples shown in the first three rows of Figure~\ref {fig:classification_errors}. One case that was not solved by the CNNs is shown in the last row (d) of Figure~\ref {fig:classification_errors} --- all the approaches produced false detections around urban areas. From these examples, we believe that these kinds of persistent errors must be addressed differently, by considering temporal analysis as described by~\cite{SCHROEDER2016210}, since it is very unlikely that an active fire will remain active at the same location for several months.

\subsection{Discussion}

In sections~\ref {section.experiments1} and~\ref {section.experiments_manual}, we discussed particular results regarding, respectively, how well the deep learning models could approximate the results from handcrafted algorithms, and how the CNNs and handcrafted sets of conditions performed when compared to a human specialist. The main findings were that the CNNs can indeed approximate the handcrafted algorithms, even with a reduced architecture and relying on only 3 channels of sensor data. Moreover, we noted that, by combining multiple CNN outputs, it is possible to obtain better performance compared to the original sets of conditions, and the situations where the CNNs differ from the original algorithms can actually be positive, allowing more complex rules and relations, albeit with the trade-off of being hard do understand or describe by a human. 

It is worth noting that the CNN architectures we described in this paper were the ones that reported the most promising results, however,  several other variations were tested, such as architectures without batch normalization (which failed, most likely due to the vanishing gradients problem), variations with more and less convolutional layers or filters per layer (i.e.~deeper/shallower, and wider/narrower architectures). We emphasize that we did not focus on finding highly optimized CNN architectures, the feasibility of the approach being our main concern. Thus, the results we obtained could be potentially improved in a number of ways: future research could explore the use of image super-resolution~(\cite{Gargiulo2019,Ma2019}); ensembles of distinct convolutional neural network architectures~(\cite{8698456}); and spatio-temporal information~(\cite{BOSCHETTI2017347}) to differentiate between temporally-persistent fire pixels from places like factories or rooftops from deforestation wildfires. Furthermore, machine learning-driven approaches can be quickly adapted to new sensors~(\cite{MATEOGARCIA20201}) (e.g. Sentinel-2), possibly relying on transfer learning, without requiring specifically designed sets of conditions.

\section{Conclusions}~\label{section.conclusions}

In this paper, we addressed the problem of active fire detection using deep learning techniques. We introduced a new large-scale dataset, containing 146,214 image patches extracted from the Landsat-8 satellite, along with associated outputs produced by three well established handcrafted algorithms (\cite{SCHROEDER2016210},~\cite{MURPHY201678} and~\cite{doi:10.1080/17538947.2017.1391341}). A second dataset was also created, with 9,044 image patches and annotations manually produced by a human specialist. We hope these datasets can prove useful for the community, since they set a challenging target while keeping the data in a friendly format for existing machine learning tools, allowing researchers to test different architectures and approaches. Besides the datasets, we presented a study on how convolutional neural networks can be used do approximate the outputs from the considered handcrafted algorithms, as well as on how the outputs from multiple models can be combined to achieve better performance than the individual sets of conditions.


Our study showed that handcrafted sets of conditions for active fire recognition are hard to be defined, and small variations in fixed thresholds may cause false positives or negatives, on the other hand, deep learning techniques are able to encode more complex rules to improve over this aspect but at the cost of relations that are hard to understand and describe by a human.

Future work shall focus on exploring alternative methods for active fire detection over the proposed datasets, as well as better exploring the hypothesis that models trained over samples from one satellite can perform well when applied to other satellites. It is also possible to develop means of combining detection results from multiple satellites, with different responses, spatial resolutions, and capture cycles; aiming at improving dependability and producing estimates with higher resolution in both space and time.


\section*{Acknowledgment}

We gratefully acknowledge the support of NVIDIA Corporation with the donation of the Titan Xp GPU used for this research. The authors would like to thank also the research Brazilian agencies CNPq, CAPES and FAPESP.

\bibliographystyle{unsrtnat}






\end{document}